\documentclass{article}
\usepackage{iclr2026_conference,times}

\usepackage{adjustbox}
\usepackage{times}
\usepackage{tabularx}
\usepackage{amssymb}
\usepackage{latexsym}
\usepackage{wrapfig}
\usepackage{longtable}
\usepackage{hyperref}   
\usepackage{booktabs}
\usepackage{tabularx}
\usepackage{multirow}

\usepackage{amsmath}
\usepackage{tcolorbox}
\tcbuselibrary{listings,skins,breakable}
\usepackage{xcolor}
\usepackage{booktabs}
\usepackage{multirow}
\usepackage{subcaption}
\usepackage[table]{xcolor}
\usepackage{verbatim}
\usepackage[T1]{fontenc}
\usepackage{enumitem}

\usepackage[utf8]{inputenc}

\usepackage{microtype}

\usepackage{inconsolata}

\usepackage{graphicx}

\author{
  \textbf{Haochen Luo\textsuperscript{1}},
  \textbf{Zhengzhao Lai\textsuperscript{2}},
  \textbf{Junjie Xu\textsuperscript{4}},
  \textbf{Yifan Li\textsuperscript{1}},
  \textbf{Tang Pok Hin\textsuperscript{1}}, 
  \textbf{Yuan Zhang\textsuperscript{3}},\\
  \textbf{Chen Liu\textsuperscript{1}}\thanks{Corresponding author: \texttt{chen.liu@cityu.edu.hk}}
\\   
  \textsuperscript{1}City University of Hong Kong \quad
  \textsuperscript{2}The Chinese University of Hong Kong (Shenzhen)\\
  \textsuperscript{3}Shanghai University of Finance and Economics \quad
  \textsuperscript{4}University of Science and Technology of China
}

\title{From Natural Language to Executable Option Strategies via Large Language Models}

\usepackage{tikz}
\usetikzlibrary{trees, positioning, arrows.meta}

\iclrfinalcopy

\begin{document}
    \maketitle
\begin{abstract}
Large Language Models (LLMs) excel at general code generation, yet translating natural-language trading intents into correct option strategies remains challenging. Real-world option design requires reasoning over massive, multi-dimensional option chain data with strict constraints, which often overwhelms direct generation methods. We introduce the \textbf{Option Query Language (OQL)}, a domain-specific intermediate representation that abstracts option markets into high-level primitives under grammatical rules, enabling LLMs to function as reliable semantic parsers rather than free-form programmers. OQL queries are then validated and executed deterministically by an engine to instantiate executable strategies. We also present a new dataset for this task and demonstrate that our neuro-symbolic pipeline significantly improves execution accuracy and logical consistency over direct baselines.
\end{abstract}

\section{Introduction}

In recent years, Large Language Models (LLMs) have demonstrated remarkable capabilities in the financial domain. While extensive research has focused on equity markets, including stock price prediction ~\citep{chen2024stock,koa2024learning}, alpha mining ~\citep{tang2025alphaagent,wang2025alpha,shi2025navigating,luo2026evoalpha}, financial news sentiment analysis and trading ~\citep{araci2019finbert, liu2025multi, xiao2025tradingagents}, the application of LLMs to \textit{financial derivatives}, particularly options, remains largely unexplored. Options are fundamental financial instruments, essential for both speculative leverage and sophisticated risk management~\citep{vine2011options}. Existing studies on machine learning in this field mainly focus on option pricing~\citep{culkin2017machine,de2018machine,ivașcu2021option,zalani2025low} and hedging~\citep{banka2025deltahedge}. By contrast, designing and executing option trading strategies has been considered the exclusive domain of quantitative analysts using rigid, programmatic trading systems. In this context, the customers provide the investment goals in Natural Language (NL) (e.g., \textit{"Find a delta-neutral Iron Condor on SPY with low implied volatility rank"}) and the quantitative analysts use their expertise to come up with trading strategies.

The rapid development of intelligent trading systems, especially with the adoption of LLMs, enables the translation of vague natural language intents into \textit{rigorous and structured} trading logic, which is a key step toward automated option trading. However, this translation remains challenging. First, traditional Natural Language Understanding models are too limited to interpret complex financial concepts such as ``Delta-neutral'' or ``Volatility Skew.'' Second, although modern LLMs possess strong domain knowledge, directly applying them to raw option chain data is impractical. Option chains contain thousands of contracts across strikes and expiries, resulting in high-dimensional inputs that exceed LLM context limits or incur high computational costs. Moreover, direct text-to-code generation often leads to hallucinations~\citep{agarwal2024codemirage}, such as producing invalid tickers or violating strict constraints, which is unacceptable in high-stakes financial settings.

\lstdefinelanguage{OQL}{
  morekeywords={SELECT,FROM,WHERE,AND,HAVING,ORDER,BY,LIMIT,DESC},
  sensitive=true,
  morecomment=[l]{--}
}

\lstset{
  language=OQL,
  basicstyle=\ttfamily\small,
  keywordstyle=\color{blue}\bfseries,
  commentstyle=\color{gray}\itshape,
  frame=single,
  breaklines=true,
  showstringspaces=false,
  numbers=none
}

To bridge this gap, we propose the \textbf{Option Query Language (OQL)}, a domain-specific intermediate representation for reliable interaction between LLMs and option markets. Instead of requiring LLMs to process massive raw data or generate error-prone Python or SQL code, our framework treats the LLM as a semantic parser that converts natural language queries into concise, syntactically constrained OQL instructions. These instructions are then deterministically executed by a dedicated compiler over the option chain. This neuro-symbolic design reduces context explosion by abstracting market data into high-level primitives and ensures logical validity through grammatical constraints.




To the best of our knowledge, this work is the first systematic effort to enable Natural Language $\rightarrow$ Option Strategy translation using LLMs. Our main contributions are as follows: 
(1) We introduce \textbf{OQL}, a domain-specific query language that encodes complex derivatives logic (e.g., Greeks, multi-leg structures, and expiry/strike relations) into a token-efficient yet execution-rigorous format, allowing LLMs to generate reliable option strategies. 
(2) We present the first benchmark for this task, including a dataset of 200 diverse option trading instructions and an evaluation suite that enables fair comparison across LLMs and baseline methods. 
(3) We extend the Text-to-SQL paradigm to a financial setting by designing a customized, SQL-like language for option strategy search, demonstrating that structured querying substantially improves reliability and end-to-end performance.




\section{Related Work}
\subsection{LLMs in Finance}

The application of Large Language Models (LLMs) in finance has evolved rapidly from textual analysis to more agentic decision-making. Early works primarily relied on Pre-trained Language Models, such as FinBERT, for sentiment analysis on financial texts~\citep{liu2021finbert}. With the advent of generative models, research attention shifted toward finance-specific foundation models, including BloombergGPT~\citep{wu2023bloomberggpt} and FinGPT~\citep{liu2023fingpt}, enabling more complicated reasoning over financial data. Building on these models, a growing number of work explores LLMs for market prediction and various trading tasks. One major application is price and return prediction by incorporating textual information such as financial news~\citep{chen2024stock,guo-hauptmann-2024-fine,wang-etal-2024-llmfactor,koa2024learning}. Related studies further investigate the capability of LLMs in searching and discovering trading signals, particularly alpha factors~\citep{li-etal-2024-large-language,shi2025navigating,luo2025efs}. More recently, several works have proposed autonomous trading agents that use LLMs as ``traders'' to rebalance portfolios based on market sentiment and reasoning~\citep{yu2024fincon,yang2025twinmarket,yu2025finmem,li2025investorbench}. To evaluate these capabilities, a number of benchmarks and datasets have been introduced, covering general financial knowledge~\citep{xie2023pixiu,xie2024finben,nie2025cfinbench}, numerical reasoning~\citep{chen2021finqa}, trading ability~\citep{li2025investorbench}, signal mining~\citep{anonymous2025alphabench}, and more complex financial tasks~\citep{zhang-etal-2025-xfinbench}. However, most existing methods and benchmarks focus on \textit{equities}. In contrast, the LLM's capability in both strategy design and evaluation to deal with trading \textit{financial derivatives}, particularly options, remains largely underexplored.


\subsection{LLMs for Text-to-Query Generation}

Translating natural-language user intent into executable query logic is commonly formulated as a semantic parsing problem, with Text-to-SQL as a representative task. Benefiting from strong code-generation capabilities, LLMs have achieved strong performance in this setting, as shown by recent benchmarks across diverse domains~\citep{hong2025next,lee2022ehrsql,gao2023text,zhang2024benchmarking}. Beyond standard SQL, recent studies further extend LLM-based query generation to specialized tasks, including query optimization~\citep{tan2025can} and edited or augmented SQL-like grammars for domain-specific logic such as multi-model query~\citep{shi-etal-2025-adaptive}, motivating our OQL framework for option strategy querying.

\section{Preliminaries}

\subsection{Properties of Option Product}

An option\footnote{We consider American options in this work. The European options may only be exercised on expiry, our proposed method can deal with them in a similar way.} is a derivative contract that grants the holder the right, but not the obligation, to buy or sell an underlying asset at a predetermined price (the \textit{strike price}) on or before a specific date (the \textit{expiration date}). These contracts are categorized as calls (the right to buy) or puts (the right to sell)The market price of an option reflects the market's expectation of future volatility, the risk and the reward profile associated with the underlying asset.

\textbf{Trading Action Space.} In option trading, the fundamental action space includes \textit{buy} or \textit{sell} decisions for both call and put options with different strikes and maturities. A single trading strategy can be expressed as a combination of such elementary actions:
\begin{equation}
a_t = \{ (d_i, s_i, K_i, T_i, q_i, p_i) \}_{i=1}^{N},
\label{eq:option}
\end{equation}
where $d_i \in \{+1, -1\}$ denotes the trade direction (\(+1\) for buying and \(-1\) for selling), $s_i \in \{\text{call}, \text{put}\}$ the option type, $K_i$ the strike price, $T_i$ the expiry time, $q_i$ the trading quantity and $p_i$ the option premium (i.e., the price paid or received when open position). The reward or payoff at maturity depends on the realized price $S_T$ of the underlying asset.

\textbf{Option Pricing Model.} The theoretical value of an option is commonly derived from stochastic models of the underlying asset price. The classical \textit{Black–Scholes–Merton (BSM)} \cite{black1973pricing} model assumes that the price $S_t$ of a non-dividend paying underlying asset at time $t$ follows a geometric Brownian motion $dS_t = \mu S_t dt + \sigma S_t dW_t$,where $\mu$ is the drift, $\sigma$ is the volatility, and $W_t$ denotes a Wiener process. Under risk-neutral valuation, the price of European call ($C$) and put ($P$) options are $C = S_tN(d_+)  - K e^{-r(T - t)}N(d_-)$ and $P = Ke^{-r(T-t)}N(-d_-)-S_tN(-d_+)$ respectively,
where $V\in \{C,P\}$ denotes a generic option price and:

\begin{equation}
d_+=\frac{\ln\!\left(\frac{S_t}{K}\right)+(r+\tfrac12\sigma^2)(T-t)}{\sigma\sqrt{T-t}},
\quad
d_-=d_+-\sigma\sqrt{T-t}.
\label{eq:option_q}
\end{equation}

\begin{wraptable}{r}{0.58\linewidth}
\vspace{-\baselineskip} 
\centering
\caption{Key option Greeks and their definitions}
\renewcommand{\arraystretch}{1.4}
\small

\begin{adjustbox}{width=\linewidth,center}
\begin{tabular}{p{0.13\linewidth} p{0.75\linewidth}}
\hline
\textbf{Greek} & \textbf{Definition} \\
\hline
\textbf{Delta} & Sensitivity of option price to changes in the underlying price, $\Delta = \frac{\partial V}{\partial S}$. \\
\textbf{Gamma} & Rate of change of Delta, $\Gamma = \frac{\partial^2 V}{\partial S^2}$. \\
\textbf{Vega}  & Sensitivity to volatility, $\nu = \frac{\partial V}{\partial \sigma}$. \\
\textbf{Theta} & Sensitivity to time decay, $\Theta = \frac{\partial V}{\partial t}$. \\
\textbf{Rho}   & Sensitivity to the risk-free interest rate, $\rho = \frac{\partial V}{\partial r}$. \\
\hline
\end{tabular}
\end{adjustbox}

\label{tab:greeks}
\vspace{-0.8em}
\end{wraptable}
Here $K$ is the strike, $T$ is the expiry date, $r$ is the risk-free rate and $N(\cdot)$ is the standard Gaussian cumulative distribution function (CDF). The sensitivities of $V$ to these underlying factors are captured by the \textit{Greeks} in Table~\ref{tab:greeks}. Because future realized volatility is unobservable, true Vega cannot be measured directly. In practice, option pricing relies on implied volatility (IV), inferred from market prices using pricing models~\cite{hull2016options}. This creates a gap between theory and real-world execution, complicating robust option strategy design.

\subsection{Option Strategy and Payoff}

The terminal payoff of a single option $(d_i, s_i, K_i, T_i, q_i, p_i)$ as defined in (\ref{eq:option}) is calculated by the equations below where $S_T$ is the underlying price at maturity $T$.
\[
P_i(S_T)=
\begin{cases}
\max(S_T-K_i,0), & s_i=\text{call},\\[3pt]
\max(K_i-S_T,0), & s_i=\text{put}.
\end{cases}
\]
Therefore, the total return of the strategy as defined in (\ref{eq:option}) is the sum over legs subtracting the option premium: $\Pi(S_T)=\sum_{i=1}^{N} q_i\, d_i \,\bigl(P_i(S_T)-p_i\bigr)$. An \textit{option strategy} is a multi-leg position formed by combining several call/put contracts on the same (or closely related) underlying assets, to achieve a target trading intent such as hedging, directional exposure, volatility trading, or income generation. Figure~\ref{fig:taxonomy} summarizes our strategy universe as a hierarchical taxonomy grouped by three high-level intents: \emph{Directional}, \emph{Volatility}, and \emph{Income \& Hedging} \cite{hull2016options}. It also reflects structural complexity via the number of legs, ranging from single-leg positions to common spreads and multi-leg structures (e.g., butterflies and condors). The full list is provided in Appendix~\ref{app:strategy_background}.

\subsection{Representation of Option Chain}



In practical trading and modeling, option information is represented in a tabular
\textit{option chain} containing discrete contract-level attributes.
At time $t$, the option chain $\mathcal{C}_t$ represents the collection of all
tradable option contracts for a given underlying asset and is defined over strikes,
maturities, and contract types:
$
\mathcal{C}_t
=
\bigl\{
c_t(K, T, s)
\;\big|\;
K \in \mathcal{K},\;
T \in \mathcal{T},\; 
s \in \{\text{call}, \text{put}\}
\bigr\}.
$
Here, $K$ denotes the strike price, $T$ the time to expiry, and $s$ the option type.
Each option contract $c_t(K, T, s)$ is associated with a set of market-observed and
model-derived attributes:
\begin{equation}
\begin{aligned}
c_t(K, T, s)
=
\bigl\{
& p_t(K, T, s),\;
v_t(K, T, s), \\
& \Delta_t(K, T, s),\;
\Gamma_t(K, T, s), \\
& \nu_t(K, T, s),\;
\Theta_t(K, T, s)
\bigr\},
\end{aligned}
\end{equation}

\usetikzlibrary{trees,arrows.meta,positioning}

\begin{wrapfigure}{r}{0.55\linewidth}
  \centering
  \vspace{-0.5em}

  \begin{adjustbox}{width=\linewidth,center}
    \begin{tikzpicture}[
      font=\sffamily,
      level 1/.style={sibling distance=5.7cm, level distance=1.8cm},
      level 2/.style={sibling distance=3cm, level distance=1.5cm},
      level 3/.style={sibling distance=2.0cm, level distance=1.5cm},
      edge from parent/.style={draw, -Latex, thick},
      every node/.style={align=center, inner sep=3pt},
      root/.style={font=\large\bfseries, text depth=0.2ex, path picture={
        \draw[thick] (path picture bounding box.south west) -- (path picture bounding box.south east);
      }},
      title/.style={font=\bfseries}
    ]

    \node[root] {Option Strategy Universe}
      child {node {Directional\\Strategies}
        child {node {Bullish}
          child {node {Long Call\\Bull Spread}}
        }
        child {node {Bearish}
          child {node {Long Put\\Bear Spread}}
        }
      }
      child {node {Volatility\\Strategies}
        child {node {Long Volatility\\(Breakout)}
          child {node {Straddle\\Strangle}}
        }
        child {node {Short Volatility\\(Range-bound)}
          child {node {Iron Condor\\Butterfly}}
        }
      }
      child {node {Income \&\\Hedging}
        child {node {Yield Gen}
          child {node {Covered Call\\Cash-Secured Put}}
        }
        child {node {Protection}
          child {node {Collar\\Protective Put}}
        }
      };
    \end{tikzpicture}
  \end{adjustbox}

  \vspace{-2mm}
  \caption{Taxonomy of option strategies in trading.}
  \label{fig:taxonomy}
  \vspace{-0.8em}
\end{wrapfigure}
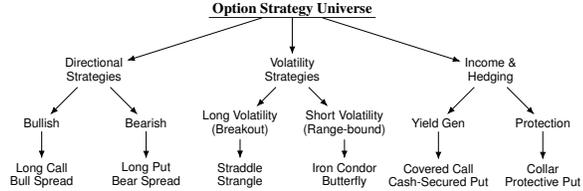

where $p_t$ and $v_t$ denote the option premium and trading volume, respectively,
and $(\Delta_t, \Gamma_t, \nu_t, \Theta_t)$ are the option Greeks defined in
Table~\ref{tab:greeks}, computed using the Black--Scholes--Merton (BSM) model. This representation provides a structured snapshot of the market state across strike,
maturity, and contract-type. Due to the combinatorial expansion of the triplet
$(K, T, s)$, the option chain at each time step is inherently high-dimensional. This scale posing a significant challenge for direct processing by LLMs.




\section{Methodology}
\label{sec:methodology}

We propose the Option Query Language (OQL) to bridge the gap between the complex, flexible natural language intent and the financial execution which requires precision.
As illustrated in Figure~\ref{fig:placeholder}, our system operates via a two-stage process: (1) \textbf{Semantic Parsing}, where an LLM translates a user's natural language intent $x$ into a structured OQL query $z$; and (2) \textbf{Deterministic Execution}, where a specialized engine validates and executes $z$ against massive market data to produce the final trading strategy $y$. This explicit separation allows linguistic reasoning and financial constraints to be handled independently, improving robustness, interpretability, and execution reliability for complex option strategies.



\subsection{Problem Formulation}
Let $\mathcal{D}$ represent the state space of the option market, containing real-time data for underlying assets, option chains, and derived risk metrics (the "Greeks"). The goal of option strategy search is to map a natural language instruction $x$ (e.g., \textit{"Find a delta-neutral iron condor on NVDA..."}) to an executable subset of contracts $y \subset \mathcal{D}$ satisfying specific logical constraints. Directly modeling the probability $\mathbb{P}(y|x, \mathcal{D})$ using a monolithic LLM is intractable due to the high dimensionality of $\mathcal{D}$ and the necessity for logical precision. Consequently, we introduce OQL as a latent intermediate representation $z$ and decompose the problem into:
\begin{equation}
    \mathbb{P}(y|x, \mathcal{D}) = \sum_{z} \mathbb{P}_{\theta}(z|x) \cdot \mathbb{P}_{\phi}(y|z, \mathcal{D})
\end{equation}
Here, $\mathbb{P}_{\theta}(z|x)$ represents the semantic parser based on an LLM parameterized by $\theta$, and $\mathbb{P}_{\phi}(y|z, \mathcal{D})$ is the deterministic compiler parameterized by $\phi$. This decoupling confines the LLM's probabilistic nature to intent parsing, while ensuring the financial execution remains verifiable and logical.

\subsection{Option Query Language (OQL)}
\label{sec:oql_design}

OQL is a declarative domain-specific language designed to represent option strategies as structured symbolic queries. Rather than enumerating procedural steps, OQL specifies structural and quantitative constraints over strategy components.

\subsubsection{Principle 1: Role-Based Abstraction}

Each feasible option strategy satisfying the user's intent $s \in \mathcal{S}$ is associated with a fixed role schema
$\mathcal{R}(s) = \{ r_1, r_2, \dots, r_k \},$
where each role $r_i$ corresponds to a semantically distinct leg in the strategy (e.g., Short Call, Long Put for Risk Reversal). A valid strategy instance $y$ must satisfy a one-to-one assignment between roles and option contracts:
$
y = \{ (r_i, c_i) \mid r_i \in \mathcal{R}(s),\ c_i \in \mathcal{D} \}.
$ 
This design enforces structural validity by construction and prevents semantically invalid combinations, such as assigning two calls to a straddle strategy. Moreover, role-level abstraction enables fine-grained constraint specification on individual legs. The details are shown in Table~\ref{tab:roles}.

\subsubsection{Principle 2: Scoped Filtering}

OQL distinguishes between constraints applied at different semantic scopes.

\noindent \textbf{Leg-level constraints} are expressed in the \texttt{WHERE} clause and operate on individual option contracts prior to strategy assembly. Formally, for each role $r$, a candidate set is defined as:
$
\mathcal{C}_r = \{ c \in \mathcal{D} \mid \psi_r(c) = \text{true} \},
$
where $\psi_r$ denotes role-specific predicates (e.g., moneyness, delta, time-to-expiry).

\noindent \textbf{Strategy constraints} are expressed in the \texttt{HAVING} clause and are applied after assembling candidate strategies. These constraints operate on aggregated properties: $\Psi(y) = \text{true}$,
where $\Psi$ may involve net Greeks, maximum loss, or reward-to-risk ratios.

\subsubsection{Principle 3: Semantic Soft-Matching}

Natural language intents often specify approximate numerical conditions. To bridge linguistic ambiguity and strict database filtering, OQL introduces an approximate matching operator $\sim$. Given a numerical attribute $a(c)$ and a target value $\tau$, the condition $a(c) \sim \tau$ is interpreted as: $|a(c) - \tau| \le \epsilon \cdot \tau,$
where $\epsilon$ is a predefined tolerance. This operator relaxes hard constraints and improves robustness by reducing empty-result failures during searching.

\begin{figure*}[h]
    \centering
    \includegraphics[width=\linewidth]{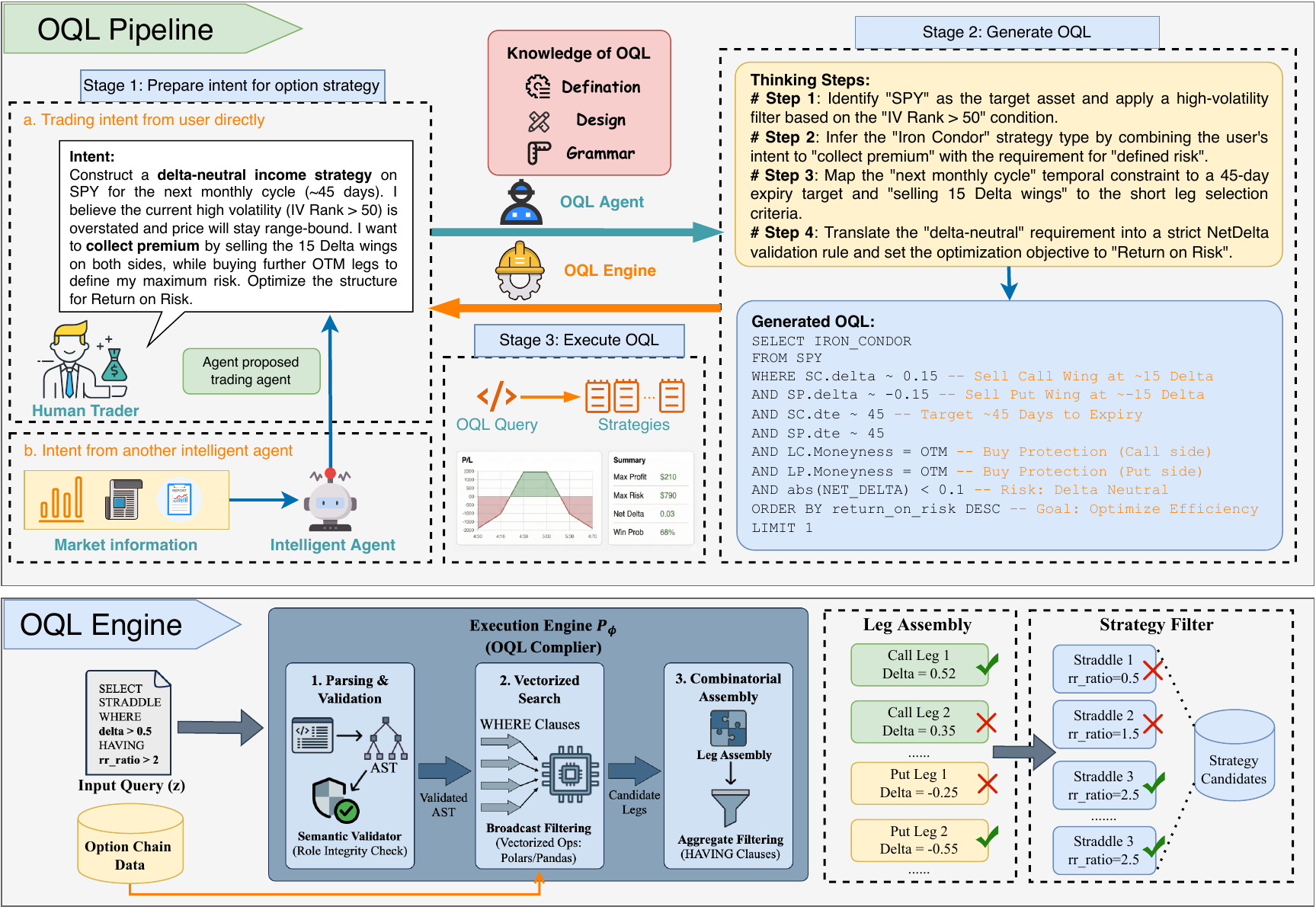}
    \caption{This figure illustrates the complete workflow of the Options Query Language (OQL) system, from intent to executable option strategies. \textbf{Top:} the OQL pipeline collects trading intent from human users or intelligent agents, translates high-level intent into formal OQL queries, and executes them to retrieve candidate strategies. \textbf{Bottom:} the deterministic OQL compiler $P_{\phi}$ processes each query through parsing and semantic validation, vectorized filtering over option-chain data, and combinatorial leg assembly with aggregate constraints, producing a ranked set of valid option strategies. The full OQL grammar, formal definitions and backend are provided in Appendix~\ref{app:oql}.}
    \label{fig:placeholder}
        \vspace{-1em}
\end{figure*}

\subsection{Neuro-Symbolic Execution Flow}

As shown in Figure~\ref{fig:placeholder}, given an OQL query $z$, the execution engine evaluates and executes $P_{\phi}(y \mid z, \mathcal{D})$ through a fully deterministic pipeline. Conceptually, this process can be viewed as constraint parsing and resolution, followed by backend-executable query construction, where the high-level OQL specification is translated into concrete operations over option-chain data. Specifically, (1) the query is first parsed into an abstract syntax tree (AST), and structural constraints induced by the role schema $\mathcal{R}(s)$ are verified to ensure semantic consistency; 
(2) leg-level predicates are applied to $\mathcal{D}$ to obtain role-specific candidate sets $\{\mathcal{C}_r\}$, implemented via vectorized filtering over the option chain; 
(3) candidate strategies are constructed through Cartesian products over $\{\mathcal{C}_r\}$, followed by strategy-level constraint evaluation, where only strategies satisfying all \texttt{HAVING} predicates are retained; and 
(4) the parsed constraints are translated into executable backend queries and executed to produce the final strategy set. Implementation details are provided in Appendix~\ref{app:oql}.

\section{Experiments}

In this section, we conduct extensive experiments to address the following three research questions.  \textbf{RQ1}: Can LLMs, under the OQL-based interaction paradigm, successfully search for executable option strategies (i.e., generate valid OQL queries that return non-empty and correct strategies)? \textbf{RQ2}: How capable are different Large Language Models (LLMs) at generating OQL queries, and how do they compare in terms of validity, accuracy, and generation quality?  \textbf{RQ3}: Do option strategies derived from OQL queries provide practical value, including outperforming baselines that give LLMs raw option data directly, and how do these searched strategies perform in backtests?

\subsection{Dataset Construction.}
To the best of our knowledge, natural-language-to-option-strategy retrieval is a relatively new task and lacks a standardized benchmark. We therefore introduce a new dataset to support this problem setting. To mitigate look-ahead bias (i.e., the “time-travel” effect where an LLM may exploit latent knowledge from pre-training)~~\citep{golchin2024time,li2025time}, we strictly restrict all market observations to 2025. We select a diverse set of underlying assets: SPY, NVDA, AAPL, GOOG, and TSLA, which exhibit distinct trend patterns and volatility regimes over the year, enabling coverage of heterogeneous market conditions.

We further partition each underlying’s 2025 price trajectory into labeled regions characterized by different movement styles (e.g., trending, reversing, range-bound, high-volatility). Both region labeling and strategy-type annotation are curated by human domain experts, who assign the most suitable strategy family for each region to ensure domain correctness. At the start of each labeled region, we write a natural-language trading intent that describes the market context and trading objective, while simulating different trader proficiency levels to reflect realistic query styles. We introduce the details and provide examples in Appendix~\ref{sec:dataset}.


\subsection{Evaluation Metrics}
\label{sec:eval_metrics}
We evaluate OQL from two complementary perspectives: \emph{query-level performance} and \emph{strategy-level performance}. Query-level metrics measure an LLM’s ability to generate valid, accurate, and semantically faithful OQL queries from natural language intent, while strategy-level metrics assess the quality of the retrieved option strategies and whether their backtesting outcomes align with the intended trading objectives. Detailed definitions and computation protocols are provided in Appendix~\ref{sec:metrics}.

\textbf{Query Quality}: To address \textbf{RQ1} and \textbf{RQ2}, we adopt a hierarchical evaluation of query quality that captures progressively stronger notions of correctness. At the most basic level, \emph{Validity Rate (VR)} measures whether generated OQL queries are syntactically well-formed and executable, i.e., they can be successfully parsed and return at least one candidate strategy. Beyond syntactic validity, \emph{Strategy Match (SM)} evaluates whether the query selects the correct option strategy family (e.g., spreads, condors) consistent with the user’s stated intent. Finally, \emph{Semantic Accuracy (SA)} assesses whether the query constraints faithfully encode the key conditions expressed in the intent—such as strikes, days-to-expiration, or Greek exposure—without omitting critical requirements or introducing unintended ones.

\textbf{Strategy Quality}: For strategy-level evaluation, we backtest the option strategies returned by each executable query and assess their performance from profitability and risk perspectives. Strategy effectiveness is measured by the \emph{Win Rate (WR)}, defined as the proportion of strategies that achieve positive end-of-period profit and loss (PnL). Risk exposure is captured by whether a strategy triggers a margin call during the backtest, reflecting its vulnerability to extreme downside scenarios. Profitability is further quantified using both the \emph{Average Profit}, defined as the mean terminal PnL across strategies, and the \emph{Return on Cost (ROC)}, computed as $\mathrm{ROC}=\mathrm{PnL}_{\mathrm{end}} / |\mathrm{cost}_{0}|$, which normalizes returns by initial capital commitment.




\subsection{Experiments  Settings.}

\textbf{Baseline settings:} We design three baselines: \textbf{Free-Form Leg Generation (FFLG)}, \textbf{Partial-Chain Grounded (PCG)}, and \textbf{Text-to-SQL}. FFLG directly prompts the LLM to generate option legs (e.g., expiry, strike, call/put, long/short, and position size) purely from natural-language intent, without access to option-chain evidence. PCG instead provides a partial option chain as structured context, grounding the generated legs in observed market data. The \textbf{Text-to-SQL} baseline translates user intent into SQL queries over predefined option-chain tables or views, relying on fixed schemas and handcrafted aggregations rather than compositional strategy reasoning. Detailed designs and implementation details of all baselines are provided in Appendix~\ref{sec:baselines}.

\textbf{Model settings:}
We evaluate a range of large language models, including commercial models such as Gemini-2.5-Flash, DeepSeek-V3 ~\citep{deepseekai2025deepseekv3technicalreport}, GPT-4.1, and GPT-4.1-Mini ~\citep{openai2024gpt4technicalreport}. We also include smaller open-weight models, such as LLaMA-3.1-8B and Qwen-3 (8B and 4B) ~\citep{yang2025qwen3technicalreport}, as well as coder-focused variants from DeepSeek and Qwen. All models use default temperature settings, and we additionally test our method under Chain-of-Thought prompting ~\citep{wei2022chain}.

\subsection{Results and Analysis}

For OQL capability and model specialization, Tables~\ref{tab:model_query_quality_by_params} and \ref{tab:model_strategy_performance_metrics_first} demonstrate that OQL effectively bridges natural language intent with executable financial logic. All large models achieved Validity Rates (VR) exceeding \textbf{0.870}, confirming the framework's robustness. A key finding is the efficiency of specialized coding models: notably, the smaller \textbf{DeepSeek-Coder-6.7B} outperforms the larger GPT-4.1-Mini in both profitability and win rate, suggesting that domain-specific syntax reasoning is more critical than pure parameter size for this task. Furthermore, we observe a trade-off between aggression and stability: while \textbf{Gemini-2.5-Flash} maximizes total Profit and ROC, \textbf{DeepSeek-Chat} offers the most risk-averse profile with the highest Win Rate and lowest tail risk.


\begin{wraptable}{r}{0.45\linewidth}
\vspace{-\baselineskip} 
\centering
\setlength{\tabcolsep}{3pt}
\renewcommand{\arraystretch}{1.05}
\small
\caption{Comparison of models on query quality. Arrows indicate optimization direction.}
\vspace{-0.5em}

\begin{adjustbox}{width=\linewidth}
\begin{tabular}{llccc}
\toprule
\textbf{Size} & \textbf{Model}
& \textbf{VR} $\uparrow$
& \textbf{SM} $\uparrow$
& \textbf{SA} $\uparrow$ \\
\midrule
\multirow{4}{*}{\textbf{Large}}
  & DeepSeek-V3          & 0.870 & \textbf{0.822} & 0.664 \\
  & Gemini-2.5 Flash     & 0.875 & 0.743          & 0.606 \\
  & GPT-4.1              & 0.935 & 0.770          & \textbf{0.698} \\
  & GPT-4.1-Mini         & \textbf{0.950} & 0.721 & 0.605 \\
\midrule
\multirow{5}{*}{\textbf{Small}}
  & LLaMA-3.1-8B         & \textbf{0.920} & 0.582 & 0.432 \\
  & Qwen2.5-Coder-7B     & 0.760 & \textbf{0.763} & 0.553 \\
  & DeepSeek-Coder-6.7B  & 0.660 & 0.659          & 0.545 \\
  & Qwen3-4B             & 0.715 & 0.671          & 0.476 \\
  & Qwen3-8B             & 0.780 & 0.808          & \textbf{0.593} \\
\bottomrule
\end{tabular}
\end{adjustbox}

\label{tab:model_query_quality_by_params}
\vspace{-1em}
\end{wraptable}

Table~\ref{tab:main_results} highlights that OQL consistently outperforms unstructured baselines (FFLG, PCG) and standard Text-to-SQL approaches. The primary advantage of OQL lies in risk management and reliability. By enforcing a structured intermediate representation, OQL significantly reduces dangerous hallucinations common in raw SQL generation. For instance, DeepSeek-Chat using OQL reduces the buyer-side Risk@90 to \textbf{18.6\%} (compared to 46.1\% with SQL) while achieving the highest overall Win Rate (\textbf{60.9\%}). This confirms that OQL's constrained search space allows models to reason more effectively about financial constraints, producing consistent alpha rather than the high-variance, high-risk outliers observed in PCG methods.

Efficiency analysis in Table~\ref{tab:token_comparison} reveals that OQL strikes an optimal balance between token consumption and retrieval validity. Unlike the PCG approach, which incurs prohibitive token costs for low retrieval yields, OQL maintains moderate token usage while achieving a dominant cache hit rate of \textbf{88.5\%}. This makes it the most cost-effective framework for high-fidelity strategy retrieval. The asset-level analysis in Appendix~\ref{appendix:add_exp_res} reveals that OQL enables models to adapt their behavior to different market conditions rather than producing uniform or rigid strategies. Some models exhibit stronger performance on volatile, trend-driven assets, while others show greater stability on index-like or mature underlyings. Importantly, OQL exposes these differences without amplifying failure modes, indicating that the framework effectively translates each model’s latent financial reasoning into executable strategies instead of constraining them to superficial syntactic patterns.

Case studies in Appendix~\ref{appendix:case_study} illustrate OQL’s proficiency in mapping high-level user intent to structurally appropriate option strategies. In hedging scenarios, OQL consistently generates accurate inverse exposures, providing robust protection during market stress. For income-oriented objectives, the generated spread strategies exhibit stable tracking and enhanced yields. OQL effectively aligns semantic intent with market structure and option-chain constraints.

\begin{table*}[h]
  \centering
  \setlength{\tabcolsep}{2.5pt} 
  \renewcommand{\arraystretch}{1.05}
  \caption{Downstream strategy performance rearranged by metrics. We compare the average over all executed strategies (\textbf{All}) and the best strategy per case (\textbf{Top}) for each metric.}
  \small
  \adjustbox{max width=0.8\textwidth}{%
  \begin{tabular}{llcccccccccc}
    \toprule
    & & \multicolumn{2}{c}{\textbf{WR} $\uparrow$} 
    & \multicolumn{2}{c}{\textbf{RE@50} $\downarrow$} 
    & \multicolumn{2}{c}{\textbf{RE@90} $\downarrow$} 
    & \multicolumn{2}{c}{\textbf{Profit} $\uparrow$} 
    & \multicolumn{2}{c}{\textbf{ROC} $\uparrow$} \\
    \cmidrule(lr){3-4} \cmidrule(lr){5-6} \cmidrule(lr){7-8} \cmidrule(lr){9-10} \cmidrule(lr){11-12}
    \textbf{Size} & \textbf{Model}
    & \textbf{All} & \textbf{Top}
    & \textbf{All} & \textbf{Top}
    & \textbf{All} & \textbf{Top}
    & \textbf{All} & \textbf{Top}
    & \textbf{All} & \textbf{Top} \\
    \midrule
    \multirow{4}{*}{\textbf{Large}}
      & DeepSeek-V3        & 0.580 & 0.609 & 0.314 & 0.316 & 0.180 & 0.195 & 368.143 & 359.776 & 0.358 & 0.418 \\
      & Gemini-2.5-Flash   & 0.558 & 0.552 & 0.361 & 0.374 & 0.191 & 0.224 & 418.192 & 331.914 & 0.270 & 0.729 \\
      & GPT-4.1            & 0.486 & 0.475 & 0.469 & 0.475 & 0.303 & 0.339 & 272.727 & 261.203 & 0.051 & 0.124 \\
      & GPT-4.1-Mini       & 0.476 & 0.489 & 0.457 & 0.452 & 0.306 & 0.306 & 211.039 & 172.547 & 0.264 & 0.554 \\
    \midrule
    \multirow{5}{*}{\textbf{Small}}
      & LLaMA-3.1-8B       & 0.420 & 0.416 & 0.359 & 0.371 & 0.190 & 0.208 & 12.430  & 2.734   & -0.018 & 0.118 \\
      & Qwen2.5-Coder-7B   & 0.476 & 0.464 & 0.361 & 0.371 & 0.208 & 0.219 & 136.469 & 151.447 & 0.178 & 0.122 \\
      & DeepSeek-Coder-6.7B& 0.503 & 0.504 & 0.401 & 0.405 & 0.239 & 0.275 & 305.907 & 301.386 & 0.217 & 0.257 \\
      & Qwen3-4B           & 0.410 & 0.406 & 0.400 & 0.420 & 0.179 & 0.217 & 118.627 & 140.538 & -0.004 & -0.000 \\
      & Qwen3-8B           & 0.483 & 0.546 & 0.468 & 0.421 & 0.284 & 0.270 & 146.626 & 174.551 & 0.197 & 0.403 \\
    \bottomrule
  \end{tabular}
  }
  \vspace{-1em}
  \label{tab:model_strategy_performance_metrics_first}
\end{table*}

\begin{table*}[h]
\centering
\small
\setlength{\tabcolsep}{4.5pt}
\caption{Performance comparison of different strategy generation methods across base LLMs. Metrics are reported in percentage (\%) where applicable. Best results per base model are bolded.}
\label{tab:main_results}
  \adjustbox{max width=0.8\textwidth}{%
\begin{tabular}{ll ccc ccc cc}
\toprule
\multirow{2}{*}{\textbf{Base Model}} & \multirow{2}{*}{\textbf{Method}} & \multicolumn{3}{c}{\textbf{Win Rate (\%)}} & \multicolumn{3}{c}{\textbf{Risk@90 (\%)}} & \multicolumn{2}{c}{\textbf{Profitability}} \\
\cmidrule(lr){3-5} \cmidrule(lr){6-8} \cmidrule(lr){9-10}
 &  & Overall & Buyer & Seller & Buyer & Seller & Wgt. & RoC & Profit \\
\midrule
\multirow{6}{*}{DeepSeek-V3} 
 & FFLG & 54.4 & 45.0 & \textbf{77.6} & 24.2 & 24.5 & 24.3 & 0.279 & 76.9 \\
 & PCG & 44.4 & 42.3 & 52.5 & 26.8 & 25.0 & 26.5 & 0.064 & 92.2 \\
 & PCG-Full & 48.9 & 43.5 & 67.4 & 24.5 & 23.3 & 24.2 & 0.258 & 117.8 \\
 & SQL & 52.8 & 43.4 & 61.2 & 46.1 & \textbf{5.9} & 24.8 & \textbf{1.364} & 195.2 \\
 & OQL (Ours) & \textbf{60.9} & \textbf{58.5} & 66.1 & \textbf{18.6} & 21.4 & \textbf{19.5} & 0.418 & \textbf{359.8} \\
 & OQL-CoT (Ours) & 60.2 & 51.3 & 76.2 & 26.5 & 20.6 & 24.4 & 0.282 & 343.5 \\
\midrule
\multirow{6}{*}{Gemini-2.5-Flash} 
 & FFLG & 57.8 & 45.5 & 83.1 & 25.6 & 22.0 & 24.4 & 0.308 & 109.5 \\
 & PCG & 61.5 & 49.6 & 79.7 & 28.1 & 25.3 & 27.0 & 0.474 & 189.5 \\
 & PCG-Full & 59.8 & 43.8 & \textbf{88.7} & 24.2 & 23.9 & 24.1 & \textbf{1.752} & 161.9 \\
 & SQL & 52.0 & 43.6 & 67.7 & 36.8 & 3.2 & 25.1 & -0.024 & 276.2 \\
 & OQL (Ours) & 55.2 & 51.3 & 81.8 & 25.7 & \textbf{0.0} & 22.4 & 0.729 & 331.9 \\
 & OQL-CoT (Ours) & \textbf{62.6} & \textbf{57.6} & 78.6 & \textbf{21.2} & 9.5 & \textbf{18.4} & 0.822 & \textbf{449.2} \\
\bottomrule
\end{tabular}
}
    \vspace{-1em}
\end{table*}



\section{Conclusion}

In conclusion, this work introduces a neuro-symbolic pipeline for option strategy search that translates natural-language trading intents into executable and verifiable strategies through an intermediate language, OQL. By decoupling semantic parsing (LLM → OQL) from deterministic execution (OQL engine → strategy set), our approach improves reliability when handling complex derivatives logic and enables large-scale evaluation across both query quality and strategy quality. We further present, to our knowledge, the first study adapting the Text-to-SQL paradigm to a customized financial domain, where the “database” corresponds to a massive option-chain space and the “query result” is a structured strategy set. Our empirical results show that semantic accuracy is more predictive of downstream strategy performance than generation success alone, highlighting the importance of faithful constraint grounding for practical option strategy search. Future work will extend OQL beyond predefined strategy templates toward free-form leg design, enabling more flexible and expressive option constructions. We also plan to support strategy queries conditioned on existing portfolio holdings, allowing the system to adaptively generate strategies based on a user’s current positions. In addition, we will expand the strategy taxonomy and OQL operators, and incorporate more realistic trading frictions and risk controls to further improve practical applicability.



\bibliographystyle{iclr2026_conference}
\bibliography{custom}

\appendix

\section{Background of Option Strategy}
\label{app:strategy_background}

We summarize the option strategies considered in this work under two common trading scenarios: \emph{directional} views (bullish/bearish exposure) and \emph{volatility} views (breakout or range-bound). Throughout this appendix, we only consider \emph{pure option} multi-leg combinations (calls/puts) on the same underlying, and we do \emph{not} include stock-option mixed positions (e.g., covered calls, protective puts, collars). All payoffs are defined at expiration as functions of $S_T$, omitting transaction costs and premium shifts.

\subsection{Directional Strategies}

\paragraph{Long Call (Bullish).}
Buy one call with strike $K$:
\begin{equation}
\Pi_{\text{LC}}(S_T)= (S_T-K)^+.
\end{equation}

\paragraph{Long Put (Bearish).}
Buy one put with strike $K$:
\begin{equation}
\Pi_{\text{LP}}(S_T)= (K-S_T)^+.
\end{equation}

\paragraph{Bull Call Spread (Bull Spread).}
Buy a call at $K_1$ and sell a call at $K_2$ with $K_1<K_2$:
\begin{equation}
\Pi_{\text{BCS}}(S_T)= (S_T-K_1)^+ - (S_T-K_2)^+.
\end{equation}

\paragraph{Bear Put Spread (Bear Spread).}
Buy a put at $K_2$ and sell a put at $K_1$ with $K_1<K_2$:
\begin{equation}
\Pi_{\text{BPS}}(S_T)= (K_2-S_T)^+ - (K_1-S_T)^+.
\end{equation}

\subsection{Volatility Strategies}

\paragraph{Long Straddle (Long Volatility).}
Buy a call and a put at the same strike $K$:
\begin{equation}
\Pi_{\text{Straddle}}(S_T)= (S_T-K)^+ + (K-S_T)^+ = |S_T-K|.
\end{equation}

\paragraph{Long Strangle (Long Volatility).}
Buy a put at $K_1$ and a call at $K_2$ with $K_1<K_2$:
\begin{equation}
\Pi_{\text{Strangle}}(S_T)= (K_1-S_T)^+ + (S_T-K_2)^+.
\end{equation}

\paragraph{Butterfly (Short Volatility / Range-Bound).}
A standard call butterfly uses three strikes $K_1<K_2<K_3$:
\begin{equation}
\begin{aligned}
\Pi_{\text{Butterfly}}(S_T)
&= (S_T-K_1)^+ - 2(S_T-K_2)^+ \\
&\quad + (S_T-K_3)^+.
\end{aligned}
\end{equation}

\paragraph{Iron Condor (Short Volatility / Range-Bound).}
Constructed by combining a put spread and a call spread with $K_1<K_2<K_3<K_4$:
\begin{equation}
\begin{aligned}
\Pi_{\text{IC}}(S_T)
&=
\underbrace{\big[(K_2-S_T)^+ - (K_1-S_T)^+\big]}_{\text{put spread}} \\
&\quad +
\underbrace{\big[(S_T-K_3)^+ - (S_T-K_4)^+\big]}_{\text{call spread}}.
\end{aligned}
\end{equation}

\section{OQL Specification and Examples}
\label{app:oql}

This appendix provides the formal definition of the Option Query Language (OQL), including its grammatical structure, role-based schema, and practical usage examples.

\subsection{Formal Syntax (EBNF)}
To facilitate Schema-Augmented Generation, OQL is defined by a strict Extended Backus-Naur Form (EBNF) grammar. This grammar is injected into the LLM context to constrain token generation and ensure syntactic validity.

\begin{verbatim}
Query     ::= SelectClause FromClause 
              WhereClause? HavingClause?
              OrderClause? LimitClause?

SelectClause ::= "SELECT" StrategyName
StrategyName ::= "BULL_CALL_SPREAD" 
               | "IRON_CONDOR" | ...

FromClause   ::= "FROM" Underlying
Underlying   ::= [A-Z]+  -- e.g., SPY
WhereClause  ::= "WHERE" LegCondition 
                 { "AND" LegCondition }
LegCondition ::= Role "." Field Op Value

HavingClause ::= "HAVING" StratCondition 
                 { "AND" StratCondition }
StratCondition ::= Field Op Value
                 | Field "BETWEEN" Val 
                   "AND" Val

OrderClause  ::= "ORDER" "BY" OrderItem 
                 { "," OrderItem }
OrderItem    ::= Field [ "ASC" | "DESC" ]
LimitClause  ::= "LIMIT" INTEGER

Role     ::= "L" | "S" | "F" | "B" | ...
Field    ::= "Dte" | "Delta" | "Iv" | ...
Op       ::= "=" | "!=" | "<" | ">" | "~"
\end{verbatim}
\subsection{Strategy Definitions and Role Schemas}
OQL enforces structural integrity through \textit{Role Schemas}. Each strategy type implies a specific set of leg identifiers (Roles) that are valid within the \texttt{WHERE} clause. The execution engine rejects queries referencing undefined roles (e.g., referencing a "Short Put" role in a "Call Spread" strategy).

\begin{table}[h]
    \centering
    \caption{Supported Strategies and Role Definitions}
    \label{tab:roles}
\begin{adjustbox}{width=0.8\linewidth}
        \begin{tabular}{l l l}
            \toprule
            \textbf{Strategy} & \textbf{Roles} & \textbf{Description} \\
            \midrule
            \texttt{BULL\_CALL\_SPREAD} & \texttt{L}, \texttt{S} & Lower-strike Long Call + Higher-strike Short Call \\
            \texttt{BEAR\_PUT\_SPREAD} & \texttt{L}, \texttt{S} & Higher-strike Long Put + Lower-strike Short Put \\
            \texttt{CALENDAR\_CALL} & \texttt{F}, \texttt{B} & Near-term Short Call + Far-term Long Call \\
            \texttt{STRADDLE} & \texttt{C}, \texttt{P} & ATM Call + ATM Put \\
            \texttt{IRON\_CONDOR} & \texttt{SC}, \texttt{LC} & Short/Long Call Wings \\
                                  & \texttt{SP}, \texttt{LP} & Short/Long Put Wings \\
            \texttt{BUTTERFLY\_CALL} & \texttt{L1}, \texttt{S}, \texttt{L2} & Long wings + Short body \\
            \bottomrule
            
        \end{tabular}
        \end{adjustbox}
\end{table}

\subsection{Query Analysis and Examples}
The following examples demonstrate how OQL handles complex financial logic that would otherwise require verbose Python scripts.

\noindent \textbf{Example 1: The "Approximate" Operator.} Natural language often implies fuzzy constraints. The tilde operator ($\sim$) allows the LLM to express "around 30 days" without hallucinating exact dates.
\begin{quote}
\noindent \textit{Intent: "Find me an Iron Condor on QQQ expiring in about a month, where I collect at least \$100 credit."}
\end{quote}
\begin{verbatim}
SELECT IRON_CONDOR FROM QQQ
WHERE  SC.Dte ~ 30 AND LC.Dte ~ 30 
  AND  SP.Dte ~ 30 AND LP.Dte ~ 30
HAVING net_credit >= 100
ORDER BY rr_ratio DESC
\end{verbatim}

\noindent \textbf{Example 2: Iron Condor (Range-Bound Income).} 
OQL supports four-leg strategies and allows users to express range-bound views using leg-level Greek constraints and maturity conditions.

\begin{quote}
\textit{Intent: ``TSLA is likely to stay range-bound. Build an iron condor with about 30 days to expiration. I want positive theta and limited downside risk.''}
\end{quote}

\begin{verbatim}
SELECT IRON_CONDOR FROM TSLA
WHERE Dte ~ 30
AND SC.Delta < 0.20
AND LC.Delta < 0.05
AND SP.Delta > -0.20
AND LP.Delta > -0.05
HAVING net_theta > 0 AND max_loss < 500
LIMIT 10
\end{verbatim}

\subsection{Backend Design}

OQL is designed as an abstract domain-specific language that serves as a middle layer between large language models (LLMs) and the concrete strategy search process. The role of the backend engine is to translate OQL queries into executable search programs over option-chain data.

In principle, OQL is backend-agnostic. After abstracting the core components of option strategy search, such as leg selection, structural constraints, and aggregate risk computation. The execution can be implemented using different backends. For example, the search process can be realized with Python-based data processing frameworks (e.g., NumPy or Pandas) for flexible prototyping, or compiled into SQL query templates or stored procedures for efficient and scalable execution in relational databases. The choice of backend depends on system requirements such as performance, scalability, and deployment constraints.

In this work, we adopt a Python-based backend to facilitate rapid prototyping and experimental evaluation. Nevertheless, the design of OQL naturally supports compilation into SQL-based execution pipelines. In future work, we plan to migrate the backend to a fully SQL-driven implementation, leveraging relational query optimization and deterministic execution to enable large-scale, reproducible strategy search.

\section{Dataset Construction}
\label{sec:dataset}

To the best of our knowledge, translating natural-language trading intents into executable multi-leg option strategies is a relatively unexplored task, and no existing dataset explicitly aligns linguistic descriptions with structured option strategies under realistic market conditions. To address this gap, we construct a new dataset by grounding expert-written intents in carefully segmented market regimes observed in 2025.

\begin{figure*}[h] 
    \centering
    \includegraphics[width=\linewidth]{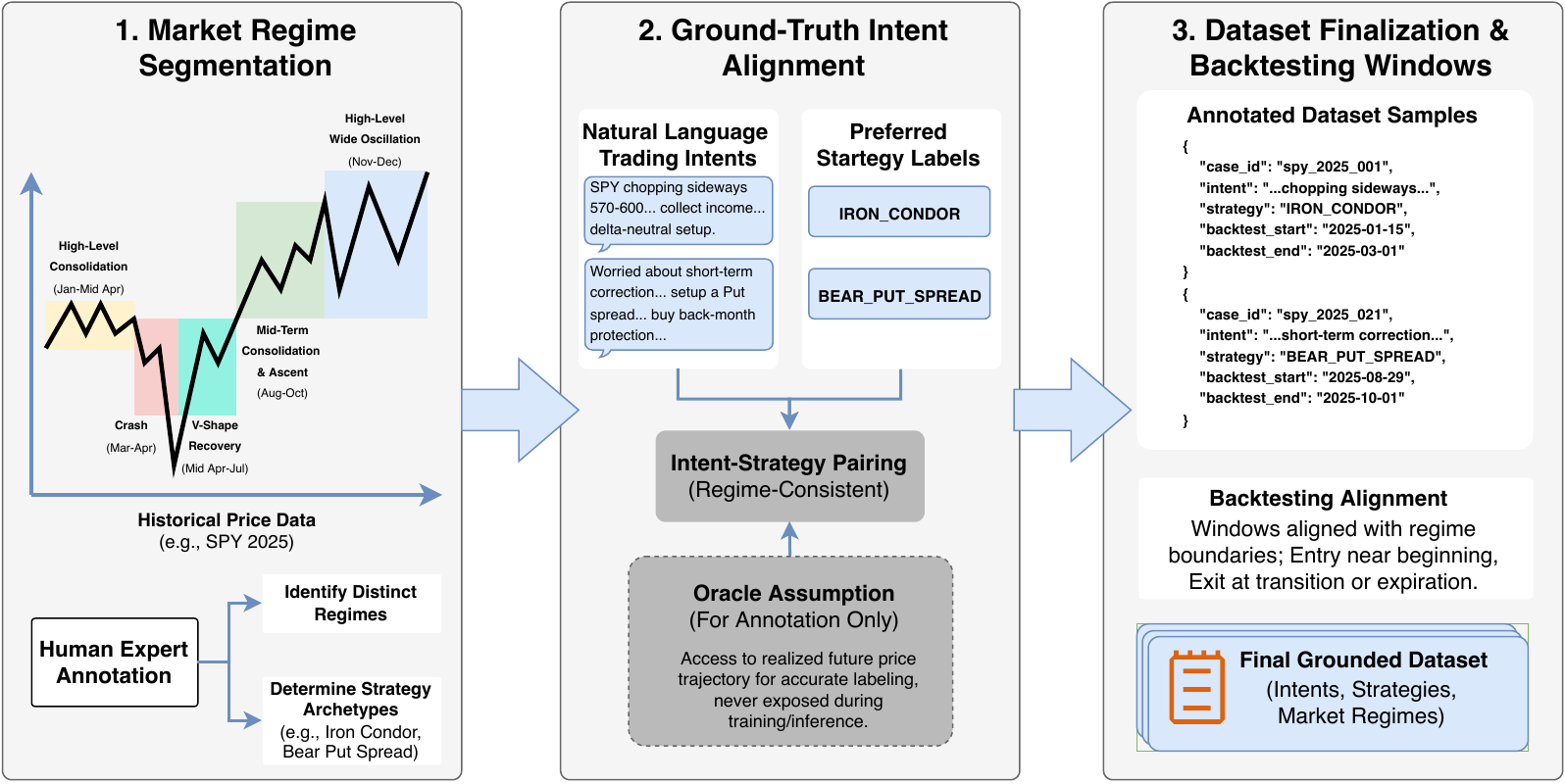}
    \caption{Dataset construction workflow. We segment historical price data into distinct market regimes, align natural-language trading intents with regime-consistent option strategy labels, and finalize JSON samples with backtesting windows aligned to regime boundaries, producing a grounded dataset of intents, strategies, and market regimes.}
    \label{fig:dataset}
\end{figure*}

In this section, we introduce the construction pipeline of our dataset. As shown in Figure~\ref{fig:dataset}, rather than relying on uniform calendar-based splits, we first identify distinct market regimes based on price action within a held-out test period that is not fully observed by large language models during training. Human experts manually classify market styles (e.g., crashes, recoveries, consolidations, or momentum-driven phases) and determine the \emph{appropriate option strategy archetypes} commonly adopted under each regime. For dataset annotation purposes, we assume a hypothetical trader who enters the market at the beginning of each regime and, as an oracle, has access to the realized price trajectory over the subsequent evaluation window. This oracle assumption is used solely to assign regime-consistent intent labels and strategy preferences, and is never exposed to the model during training or inference. The resulting regimes and strategy mappings are employed exclusively for intent grounding and evaluation, ensuring that no look-ahead information is exploited.

\subsection{Market Regime Segmentation}

\begin{table*}[h]
\centering
\small 
\caption{Two example intent cases in our benchmark. We categorize queries by user expertise (e.g., Junior vs. Senior) and extract key parameters such as the underlying asset, preferred strategy, and backtesting range from the unstructured intent description.}
\begin{tabularx}{\linewidth}{l l l X l}
\toprule
\textbf{Case Type} & \textbf{Underlying} & \textbf{Strategy} & \textbf{User Intent Description} & \textbf{Backtest Period} \\
\midrule
Example 1 & NVDA & Bull Call & ``Okay, we are bouncing off 95. Jensen is speaking next week. & 2025-04-15 \\
(Junior)  &      & Spread    & I think the bottom is in. I want to catch the run back to 120, & to \\
          &      &           & but I don't have much cash. Get me a cheap Call Spread.''      & 2025-05-15 \\
\midrule
Example 2 & GOOG & Bear Call & ``The stock is stuck below the 210 resistance and looks heavy. & 2025-01-17 \\
(Senior)  &      & Spread    & I want to sell the 215/220 call spread to collect premium.     & to \\
          &      &           & I don't think it has the energy to break out...''              & 2025-02-10 \\
\bottomrule
\end{tabularx}
\label{tab:intent_examples}
\end{table*}

For each underlying asset, we manually divide the yearly price trajectory into consecutive regimes based on dominant trend direction, volatility characteristics, drawdown magnitude, and structural price patterns (e.g., V-shaped reversals, range-bound consolidation, or breakout acceleration). Each regime represents a distinct market condition that naturally corresponds to different option strategy preferences. Below, we illustrate this process using representative examples from \textbf{SPY} in Table~\ref{tab:spy_regime_strategy},
which serves as a market-wide benchmark and exhibits a pronounced deep V-shaped reversal in 2025, followed by a sustained bull market and high-level consolidation. Based on price action, we identify four major regimes.


\begin{table*}[h]
\centering
\caption{Market regime segmentation of SPY in 2025 and corresponding option strategy preferences.}
\label{tab:spy_regime_strategy}
\small 
\setlength{\tabcolsep}{6pt} 

\begin{tabularx}{\linewidth}{p{3.6cm} >{\raggedright\arraybackslash}X >{\raggedright\arraybackslash}X}
\toprule
\textbf{Regime \& Time Period} & \textbf{Price Action Characteristics} & \textbf{Typical Option Strategy Preference} \\
\midrule
\textbf{I: High-level consolidation \& drawdown} \newline \textit{(Jan -- mid Apr)}
&
Range-bound (570--600) followed by rapid Feb sell-off. Breaks support to year-low ($\sim$500) with high volume, indicating panic liquidation.
&
Bearish hedging structures (long puts, bear spreads) to protect against tail risk and capture downside convexity.
\\
\midrule
\textbf{II: V-shaped rebound \& uptrend} \newline \textit{(mid Apr -- Jul)}
&
Double-bottom in Apr, then steep rebound. Rallies from $\sim$500 to $>$640 in 3 months with strong momentum and minimal pullbacks.
&
Risk-controlled bullish strategies (bull spreads, call diagonals) to leverage upside while managing capital and volatility.
\\
\midrule
\textbf{III: Consolidation \& stair-step advance} \newline \textit{(Aug -- Oct)}
&
Momentum moderates. Intermittent pullbacks ($\sim$620) and renewed advances form a zigzag structure with higher highs and lows.
&
Moderately bullish/volatility-selling (call spreads, calendars, directional strangles) for growth with reduced trend strength.
\\
\midrule
\textbf{IV: Volatility \& stabilization} \newline \textit{(late Oct -- Dec)}
&
Sharp correction (688 to 650) then rapid recovery. Stabilizes near highs ($\sim$683) in wide consolidation with elevated volatility.
&
Non-directional income strategies (iron condors, short straddles) to capture time decay amidst high volatility and uncertainty.
\\
\bottomrule
\end{tabularx}
\end{table*}

\subsection{Ground-Truth Intent Alignment and Backtesting Windows}

For each market regime, we construct ground-truth annotations that link observed price-action patterns to natural-language trading intents and corresponding option strategy preferences, we provide examples in Table~\ref{tab:intent_examples}. Intents are written by human experts based solely on information available up to the regime endpoint and describe high-level objectives such as directional bias (bullish, bearish, or neutral) and risk preference (e.g., downside protection, upside participation, or volatility harvesting), without referencing any future price realization. Each regime is associated with typical strategy archetypes commonly used under similar conditions (e.g., downside-protection strategies during drawdowns or non-directional income strategies during high-level consolidation), as summarized in Table~\ref{tab:spy_regime_strategy}. 

For quantitative evaluation, backtesting windows are manually aligned with regime boundaries. The entry date is set near the beginning of the corresponding regime, while the exit date is chosen at the regime transition or at option expiration, whichever occurs first. This design ensures a consistent and transparent mapping between market observations, natural-language intents, and backtested strategy outcomes.


\section{Details in Experiments}

\subsection{Evaluation Metrics}
\label{sec:metrics}

We evaluate the system from two complementary perspectives: \emph{Query Quality} (validity, efficiency, and semantic correctness of OQL generation) and \emph{Strategy Quality} (financial performance of the generated strategies).

\subsubsection{Query Quality}
To address RQ1 and RQ2, we employ a hierarchical set of metrics to evaluate the LLMs' proficiency in OQL query generation. These metrics range from syntactic validity to semantic faithfulness and execution efficiency. As we introduce three metrics to measure correctness and alignment used in Section~\ref{sec:eval_metrics}, in this section, we provide more details in Semantic Accuracy (SA). We use an LLM-based evaluator (GPT-4o) to judge whether the query's constraints correctly reflect the intent's key conditions (e.g., strikes, DTE, Greeks) without missing hard constraints or hallucinating requirements. The specific prompt used for this evaluation is detailed in Figure~\ref{fig:sa_prompt}.

We additionally design two metrics to further measure the query efficiency and selectivity. Let $\mathcal{D} = \{1, \dots, N\}$ be the set of test cases. For each case $j$, the model is allowed up to $K$ attempts. Let $k_j$ denote the index of the first successful attempt (parsable and non-empty). If no success occurs within $K$ tries, we set $k_j = \infty$.

\begin{figure*}[h]
\centering
\begin{tcolorbox}[colback=gray!5!white,colframe=gray!75!black,
    boxsep=2pt, left=2pt, right=2pt, top=2pt, bottom=2pt, 
    fontupper=\linespread{1.2}\selectfont,
    title=Prompt for Semantic Accuracy (SA) Evaluation]
\footnotesize
\textbf{System Instruction:} You are an evaluator for semantic match between a natural-language intent and an OQL query.

\textbf{Goal:}
\begin{itemize}[leftmargin=*, nosep] 
    \item Judge whether the OQL query is a reasonable and faithful translation of the intent.
    \item This is NOT a formal proof task. Allow reasonable interpretations for vague wording.
\end{itemize}

\textbf{Key Ideas:}
\begin{itemize}[leftmargin=*, nosep]
    \item \textbf{HARD Constraints:} Explicit numeric/categorical requirements (e.g., DTE=30, ATM/OTM, net\_credit $\ge$ X, max\_loss $\le$ Y).
    \item \textbf{SOFT Constraints:} Approximate phrases ("around", "near", "low risk").
\end{itemize}

\textbf{Evaluation Rules:}
\begin{enumerate}[leftmargin=*, nosep]
    \item \textit{Core Strategy Match:} FAIL if the SELECT strategy family differs from the intent's main trading idea.
    \item \textit{HARD Constraints:} FAIL if violated. PARTLY\_CORRECT if details are missing but not contradictory.
    \item \textit{SOFT Constraints:} Treat flexibly. Missing some does not automatically fail.
    \item \textit{Extra Constraints:} Do NOT penalize structural validity constraints.
    \item \textit{Approximation:} Treat "$\sim$" or "around" with tolerance (e.g., DTE $\pm$ 5).
\end{enumerate}

\textbf{Grading Scale:}
\begin{itemize}[leftmargin=*, nosep]
    \item \texttt{completely\_correct}: Core match + HARD constraints satisfied.
    \item \texttt{partly\_correct}: Core match + minor details missing or extra constraints narrowing intent.
    \item \texttt{fail}: Wrong strategy or contradiction of HARD constraints.
\end{itemize}

\textbf{Output Format:} JSON with keys \texttt{grade} and \texttt{comment}.
\end{tcolorbox}
\caption{The LLM-as-a-Judge prompt used to evaluate Semantic Accuracy (SA). It distinguishes between hard and soft constraints to ensure fair evaluation of the generated OQL code.}
\label{fig:sa_prompt}
\end{figure*}

\paragraph{Efficiency (Eff).}
This metric penalizes multiple retries. It is defined as the average remaining budget ratio for successful cases:
\begin{equation}
    \mathrm{Eff} = \frac{1}{N} \sum_{j=1}^{N} \mathbb{I}(k_j \le K) \cdot \left( 1 - \frac{k_j}{K} \right),
\end{equation}
where $\mathbb{I}(\cdot)$ is the indicator function.

\paragraph{Selectivity (AvgRows).}
To measure query specificity, we report the average number of rows ($r_{j, k_j}$) returned by successful queries. Given the constraint of \texttt{LIMIT 10}, this is calculated as:
\begin{equation}
    \mathrm{AvgRows} = \frac{1}{|\mathcal{S}|} \sum_{j \in \mathcal{S}} r_{j, k_j},
\end{equation}
where $\mathcal{S} = \{j \mid k_j \le K\}$ is the set of solved cases. Lower values indicate tighter, more specific constraints.


\subsubsection{Strategy Quality}
We conduct backtests for $M$ generated strategies. Let $c_i$ be the initial net cash flow for strategy $i$. We classify strategies as \textbf{Buyer} (Net Debit, $c_i > 0$) or \textbf{Seller} (Net Credit, $c_i < 0$). Let $P_i(t)$ denote the cumulative PnL at time $t \in [0, T]$.

\paragraph{Win Rate (WR).}
The proportion of strategies with a positive final PnL:
\begin{equation}
    \mathrm{WR} = \frac{1}{M} \sum_{i=1}^{M} \mathbb{I}(P_i(T) > 0).
\end{equation}
We report this for the overall set, as well as distinct Buyer and Seller subsets.

\paragraph{Risk Exposure (RE).}
To capture tail risk, we measure if the interim drawdown exceeds a fraction $\tau \in \{0.5, 0.9\}$ of the initial premium magnitude $|c_i|$. The metric $\text{RE}_\tau$ is defined as:
\begin{equation}
\label{eq:re_metric}
    \mathrm{RE}_\tau = \frac{1}{M} \sum_{i=1}^{M} \mathbb{I}\left( \min_{t \in [0, T]} P_i(t) \le -\tau |c_i| \right).
\end{equation}
\textbf{RE50} and \textbf{RE90} correspond to $\tau=0.5$ and $0.9$, respectively. High RE values indicate significant downside risk, which is particularly critical for Seller strategies that may otherwise show high win rates.

\subsection{Design of Baselines}
\label{sec:baselines}
\begin{table*}[h]
\centering
\small
\renewcommand{\arraystretch}{1.3}
\caption{Comparison of baseline prompting strategies and output formats. \textbf{FFLG} relies entirely on the model’s internal priors to infer strikes. \textbf{PCG} grounds generation using a partial option chain, enabling direct contract selection. \textbf{Text-to-SQL} translates intent into an executable database query that retrieves valid multi-leg strategies.}
\begin{tabularx}{\linewidth}{@{} l p{0.45\linewidth} X @{}}
\toprule
\textbf{Method} & \textbf{Input Prompt (Abbreviated)} & \textbf{Model Output (Snippet)} \\
\midrule

\textbf{FFLG} &
\textbf{System:} You are an options trading strategy expert. Given a trading intent, output a JSON strategy blueprint with no additional text. Available strategies include Iron Condor, Spreads, Straddles, and Butterflies. \newline
\textbf{User:} The stock is consolidating nicely around 240. IV is dropping. Sell an Iron Condor to capture theta decay. I don’t expect a breakout this week. Underlying: TSLA. &
\texttt{\{"strategy\_type":"IRON\_CONDOR",} \newline
\texttt{\phantom{"}"legs":\{} \newline
\texttt{\phantom{"}"SC":\{"strike":255\},} \newline
\texttt{\phantom{"}"LC":\{"strike":260\},} \newline
\texttt{\phantom{"}"SP":\{"strike":225\},} \newline
\texttt{\phantom{"}"LP":\{"strike":220\}\}\}} \\
\midrule

\textbf{PCG} &
\textbf{System:} You are an options trading strategy expert. Given a trading intent and a grounded option chain table for reference, output a JSON strategy blueprint. The generated legs must correspond to contracts in the provided chain. \newline
\textbf{Option Chain (Sample):} \newline
{\scriptsize
\texttt{| ticker | strike | type | close | delta | ... |} \newline
\texttt{| O:TSLA251219P00300000 | 300 | P | 19.95 | -0.29 | ... |} \newline
\texttt{| O:TSLA251219C00600000 | 600 | C | 2.75 | 0.06 | ... |}
} \newline
\textbf{User:} I want to hold TSLA long-term but volatility is too high. Zero-cost collar expiring next year. Cap upside at 600 and protect downside at 300. &
\texttt{\{"strategy\_type":"COLLAR",} \newline
\texttt{\phantom{"}"strategy\_details":\{} \newline
\texttt{\phantom{"}"contract\_ticker\_LP":} \newline
\texttt{\phantom{"}"O:TSLA251219P00300000",} \newline
\texttt{\phantom{"}"price\_LP":19.95,} \newline
\texttt{\phantom{"}"contract\_ticker\_SC":} \newline
\texttt{\phantom{"}"O:TSLA251219C00600000",} \newline
\texttt{\phantom{"}"price\_SC":2.75\}\}} \newline
\textit{(Contracts selected directly from provided chain)} \\
\midrule

\textbf{Text-to-SQL} &

\textbf{System / Instruction}

You are an expert in options trading and SQL. Your task is to generate \emph{one single, syntactically valid SQL query} that searches for option-strategy candidates from an options-chain database based on a natural-language trading intent.

You MUST output \emph{exactly one JSON object} that strictly follows the predefined schema (no markdown, no explanation, no extra text).


\textbf{User:} TSLA may pause around 450 before the next leg up. Construct a call butterfly centered at 450 for late November. Prefer low net debit. &
\texttt{\{"strategy\_type":"BUTTERFLY\_CALL",} \newline
\texttt{\phantom{"}"sql":"SELECT l1.ticker AS contract\_ticker\_L1,} \newline
\texttt{\phantom{"}s.ticker AS contract\_ticker\_S,} \newline
\texttt{\phantom{"}l2.ticker AS contract\_ticker\_L2} \newline
\texttt{\phantom{"}FROM tsla\_2025\_11\_10\_C l1} \newline
\texttt{\phantom{"}JOIN tsla\_2025\_11\_10\_C s ON s.strike=450} \newline
\texttt{\phantom{"}JOIN tsla\_2025\_11\_10\_C l2 ON l2.strike=460} \newline
\texttt{\phantom{"}ORDER BY net\_debit ASC LIMIT 10"\}} \\
\bottomrule
\end{tabularx}

\label{tab:baseline_comparison}
\end{table*}

To validate the necessity of a domain-specific intermediate representation (OQL), we compare our framework against three representative paradigms: direct generation, context-grounded generation, and general-purpose query generation.

\paragraph{1. Free-Form Leg Generation (FFLG).}
This baseline represents the capability of an LLM to generate option strategies relying solely on its parametric knowledge, without access to external market data. 
The model receives the natural language intent and is instructed to output a structured JSON containing the option legs (e.g., expiry, strike, type).
This baseline evaluates whether an LLM can "hallucinate" a correct strategy structure, serving as a lower bound for performance.

\paragraph{2. Partial-Chain Grounded (PCG).}
This baseline incorporates a Retrieval-Augmented Generation (RAG) approach. Alongside the user's intent, the model is provided with a snapshot of the option chain (e.g., the top-$N$ most liquid contracts around the at-the-money price).
The model must select specific contracts from this context to construct the strategy. This evaluates the model's ability to ground its reasoning in observed data, though it is limited by the context window and cannot perform complex filtering across the entire database.

\paragraph{3. Text-to-SQL (SQL).}
This baseline represents the standard industry approach for database interaction. The model is provided with the full database schema (table definitions and column descriptions) and is tasked with translating the natural language intent into a standard SQL query.
While SQL is expressive, it lacks high-level abstractions for financial logic (e.g., calculating spread costs or Greeks often requires complex joins and nested queries), making it a strong but challenging baseline for complex logical reasoning.

\section{Additional Results}
\label{appendix:add_exp_res}

\paragraph{Token Usage.} We evaluate the computational cost by measuring the token usage using DeepSeek-Chat, averaged across all cases in the dataset. The results are summarized in Table~\ref{tab:token_comparison}.

In this appendix, we provide a granular analysis of model performance, highlighting key behaviors across different asset classes and validating the effectiveness of our proposed generation frameworks compared to the baseline.

\paragraph{High-Volatility Adaptation vs. General Stability.} We observe a distinct divergence in model behavior relative to asset volatility. DeepSeek-Chat demonstrates exceptional adaptation to high-momentum assets, achieving dominant profitability on volatile tickers such as NVDA (Profit: 481.3, WR: 0.697) and TSLA (Profit: 984.0, WR: 0.800). This suggests an "aggressive" internal bias suitable for capturing large price swings. In contrast, Gemini 2.5 Flash exhibits superior stability and risk management on broader market indices and mature assets. It outperforms all peers on SPY (Profit: 552.4, ROC: 0.860) and maintains the highest win rate on AAPL (0.657). This dichotomy indicates that while DeepSeek-Chat excels in maximizing alpha in trending markets, Gemini 2.5 Flash offers a more robust baseline for general-purpose strategy construction.

\begin{wraptable}{r}{0.55\linewidth}
    \vspace{-1em}
    \centering
    \small
    \setlength{\tabcolsep}{4pt}
    \caption{Comparison of token usage and cache hit rates across different methods. Our method achieves the highest cache hit rate.}
    \label{tab:token_comparison}
    \vspace{-6pt}
    \adjustbox{max width=\textwidth}{%
    \begin{tabular}{lrrrr}
        \toprule
        & \multicolumn{3}{c}{\textbf{Tokens}} & \\
        \cmidrule(lr){2-4}
        \textbf{Method} & Prompt & Compl. & Total & \textbf{Hit Rate} \\
        \midrule
        FFLG & 405.34 & 40.60 & 445.94 & 62.1\% \\
        Text-to-SQL & 1395.02 & 260.20 & 1655.22 & 81.9\% \\
        \textbf{OQL (Ours)} & 2396.60 & 87.56 & 2484.16 & \textbf{88.5\%} \\
        PCG & 3079.33 & 127.62 & 3206.95 & 22.4\% \\
        PCG + Full Table & 7677.21 & 127.97 & 7805.18 & 8.0\% \\
        \bottomrule
    \end{tabular}
    }
    \vspace{-10pt}
\end{wraptable}

\paragraph{The Gap Between Syntax and Alpha.} A critical finding from Table~\ref{tab:model_performance_comparison_top} is that high syntactic validity does not guarantee financial viability. Specialized coding models like DeepSeek-Coder and Qwen-Coder frequently achieve competitive Efficiency scores (e.g., DeepSeek-Coder Eff: 0.773 on AAPL) but fail to translate this into consistent returns. For instance, on GOOG, Llama-3.1-8B and Qwen3-4B incurred significant losses (-144.8 and -48.8, respectively) despite acceptable validity rates. This highlights that financial reasoning, the ability to identify causal market factors is a distinct capability from merely adhering to the OQL grammar. Smaller models often "overfit" to the syntax without encoding meaningful trading logic.

\begin{table*}[h]
  \centering
  \scriptsize
  \setlength{\tabcolsep}{1.5pt} 
  \caption{Performance metrics across five underlyings: Win Rate (WR), Risk90, Profit, and ROC. (All)}
  \label{tab:model_performance_comparison_top}

  \adjustbox{max width=\textwidth}{%
  \begin{tabular}{lcccccccccccccccccccc}
    \toprule
    \multirow{2}{*}{\textbf{Model}} 
    & \multicolumn{4}{c}{\textbf{AAPL}} 
    & \multicolumn{4}{c}{\textbf{GOOG}} 
    & \multicolumn{4}{c}{\textbf{NVDA}} 
    & \multicolumn{4}{c}{\textbf{SPY}} 
    & \multicolumn{4}{c}{\textbf{TSLA}} \\
    \cmidrule(lr){2-5} \cmidrule(lr){6-9} \cmidrule(lr){10-13} \cmidrule(lr){14-17} \cmidrule(l){18-21}
    & \textbf{WR} & \textbf{Risk90} & \textbf{Profit} & \textbf{ROC}
    & \textbf{WR} & \textbf{Risk90} & \textbf{Profit} & \textbf{ROC}
    & \textbf{WR} & \textbf{Risk90} & \textbf{Profit} & \textbf{ROC}
    & \textbf{WR} & \textbf{Risk90} & \textbf{Profit} & \textbf{ROC}
    & \textbf{WR} & \textbf{Risk90} & \textbf{Profit} & \textbf{ROC} \\
    \midrule

    \multicolumn{21}{c}{\cellcolor{gray!15}\textbf{FFLG}} \\
    DeepSeek-Chat & 0.389 & 0.333 & 24.73 & -0.157 & 0.667 & 0.212 & 130.50 & 0.443 & 0.571 & 0.171 & 122.85 & 0.146 & 0.606 & 0.333 & 112.68 & 0.218 & 0.500 & 0.156 & -6.28 & 0.176 \\
    Gemini-2.5-Flash & 0.432 & 0.324 & 16.70 & -0.123 & 0.611 & 0.194 & 152.70 & 0.175 & 0.611 & 0.111 & 118.00 & 0.558 & 0.735 & 0.294 & 195.43 & 0.476 & 0.514 & 0.297 & 64.90 & 0.176 \\

    \multicolumn{21}{c}{\cellcolor{gray!15}\textbf{PCG}} \\
    DeepSeek-Chat & 0.425 & 0.350 & 160.03 & -0.245 & 0.395 & 0.237 & -60.03 & 0.023 & 0.462 & 0.154 & 356.88 & 0.149 & 0.564 & 0.231 & 397.64 & 0.622 & 0.364 & 0.364 & -458.26 & -0.392 \\
    Gemini-2.5-Flash & 0.500 & 0.375 & 158.83 & 0.012 & 0.600 & 0.200 & 138.23 & 0.273 & 0.675 & 0.175 & 330.28 & 0.327 & 0.675 & 0.350 & 488.13 & 0.533 & 0.625 & 0.250 & -167.88 & 0.459 \\

    \multicolumn{21}{c}{\cellcolor{gray!15}\textbf{PCG Full Table}} \\
    DeepSeek-Chat & 0.400 & 0.325 & 41.03 & -0.033 & 0.450 & 0.250 & 11.90 & 0.044 & 0.462 & 0.256 & 50.05 & 0.358 & 0.641 & 0.179 & 408.33 & 0.502 & 0.500 & 0.188 & 76.06 & -0.001 \\
    Gemini-2.5-Flash & 0.500 & 0.225 & 43.35 & 0.021 & 0.538 & 0.282 & 0.65 & 0.096 & 0.700 & 0.050 & 321.43 & 0.930 & 0.750 & 0.325 & 391.70 & 7.087 & 0.500 & 0.325 & 52.55 & 0.315 \\

    \multicolumn{21}{c}{\cellcolor{gray!15}\textbf{Text2SQL}} \\
    DeepSeek-Chat & 0.435 & 0.348 & 118.67 & 0.030 & 0.487 & 0.123 & 83.26 & 0.047 & 0.543 & 0.215 & 392.87 & 0.974 & 0.632 & 0.301 & 571.22 & 4.882 & 0.419 & 0.248 & 116.57 & 0.186 \\
    Gemini-2.5-Flash & 0.487 & 0.227 & 199.59 & -0.001 & 0.451 & 0.180 & 24.41 & 0.130 & 0.593 & 0.332 & 214.24 & -0.049 & 0.550 & 0.231 & 472.20 & 0.360 & 0.561 & 0.204 & 543.71 & 0.024 \\

    \bottomrule
  \end{tabular}
  }
\end{table*}

\begin{table*}[h]
  \centering
  \scriptsize
  \setlength{\tabcolsep}{1.5pt} 
  \caption{Performance metrics across five underlyings: Win Rate (WR), Risk90, Profit, and ROC. (All)}
  \label{tab:model_performance_comparison_top}

  \adjustbox{max width=\textwidth}{%
  \begin{tabular}{lcccccccccccccccccccc}
    \toprule
    \multirow{2}{*}{\textbf{Model}} 
    & \multicolumn{4}{c}{\textbf{AAPL}} 
    & \multicolumn{4}{c}{\textbf{GOOG}} 
    & \multicolumn{4}{c}{\textbf{NVDA}} 
    & \multicolumn{4}{c}{\textbf{SPY}} 
    & \multicolumn{4}{c}{\textbf{TSLA}} \\
    \cmidrule(lr){2-5} \cmidrule(lr){6-9} \cmidrule(lr){10-13} \cmidrule(lr){14-17} \cmidrule(l){18-21}
    & \textbf{WR} & \textbf{Risk90} & \textbf{Profit} & \textbf{ROC}
    & \textbf{WR} & \textbf{Risk90} & \textbf{Profit} & \textbf{ROC}
    & \textbf{WR} & \textbf{Risk90} & \textbf{Profit} & \textbf{ROC}
    & \textbf{WR} & \textbf{Risk90} & \textbf{Profit} & \textbf{ROC}
    & \textbf{WR} & \textbf{Risk90} & \textbf{Profit} & \textbf{ROC} \\
    \midrule
    DeepSeek-Chat & 0.584 & 0.111 & 214.1 & 0.223 & 0.419 & 0.151 & 7.6 & 0.204 & \textbf{0.642} & 0.163 & \textbf{486.8} & 0.646 & 0.456 & 0.338 & 83.1 & 0.242 & \textbf{0.765} & \textbf{0.161} & \textbf{1006.4} & 0.491 \\
    DeepSeek-Coder & 0.529 & 0.278 & 165.9 & 0.082 & 0.413 & 0.218 & \textbf{48.3} & \textbf{0.343} & 0.508 & 0.204 & 290.3 & 0.177 & 0.496 & 0.292 & 314.6 & 0.119 & 0.557 & 0.184 & 713.1 & 0.419 \\
    Gemini-2.5-Flash & \textbf{0.645} & 0.114 & 188.8 & \textbf{0.314} & 0.455 & 0.123 & -19.7 & 0.026 & 0.583 & 0.196 & 425.1 & 0.219 & \textbf{0.554} & 0.346 & \textbf{602.3} & 0.319 & 0.526 & 0.164 & 778.3 & 0.334 \\
    GPT-4.1 & 0.528 & 0.202 & 105.1 & -0.101 & \textbf{0.473} & 0.182 & 3.1 & -0.166 & 0.402 & 0.390 & 325.5 & \textbf{4.661} & 0.417 & 0.446 & 227.5 & 0.001 & 0.613 & 0.294 & 704.7 & 0.332 \\
    GPT-4.1-Mini & 0.489 & 0.238 & 98.5 & 0.152 & 0.417 & 0.241 & -9.9 & -0.022 & 0.426 & 0.399 & 139.1 & 0.203 & 0.477 & 0.367 & 140.2 & 0.394 & 0.556 & 0.296 & 656.2 & 0.513 \\
    Llama-3.1-8B & 0.460 & 0.254 & 68.6 & 0.026 & 0.362 & \textbf{0.096} & -126.4 & -0.134 & 0.374 & \textbf{0.114} & -39.7 & 0.047 & 0.393 & 0.263 & 20.3 & -0.079 & 0.506 & 0.215 & 134.0 & 0.052 \\
    Qwen-Coder & 0.433 & 0.205 & 161.4 & 0.123 & 0.447 & 0.116 & -30.9 & -0.030 & 0.453 & 0.177 & -123.6 & 0.067 & 0.479 & 0.311 & 158.5 & \textbf{0.621} & 0.573 & 0.236 & 505.9 & 0.445 \\
    Qwen3-4B & 0.566 & \textbf{0.090} & \textbf{290.9} & 0.153 & 0.235 & 0.173 & -117.1 & -0.240 & 0.390 & 0.248 & -25.6 & 0.015 & 0.417 & \textbf{0.221} & 73.7 & -0.120 & 0.378 & 0.203 & 302.0 & 0.097 \\
    Qwen3-8B & 0.528 & 0.208 & 182.1 & 0.125 & 0.393 & 0.167 & -156.0 & -0.157 & 0.498 & 0.236 & 292.4 & 0.115 & 0.473 & 0.473 & 89.0 & 0.375 & 0.516 & 0.361 & 335.7 & \textbf{0.581} \\
    \bottomrule
  \end{tabular}
  }
\end{table*}

\paragraph{Impact of Structured Generation (Baseline Comparison).} Table~\ref{tab:baseline_comparison} provides compelling evidence for the necessity of our structured Text2SQL framework over the unstructured FFLG baseline. The FFLG approach consistently resulted in poor risk-adjusted returns, with DeepSeek-Chat posting a negative ROC (-0.157) on AAPL and negligible profits on TSLA. By constraining the output space with OQL (Text2SQL), we unlock the models' latent reasoning capabilities. This is most visible on SPY, where DeepSeek-Chat's performance surged from a baseline ROC of 0.218 (FFLG) to an extraordinary 4.882 (Text2SQL). This massive deltas confirms that the primary bottleneck for LLM-based quant research is not the lack of financial knowledge, but the inability to articulate it executably without a formal grammar.

\paragraph{Prompting Paradigms: PCG vs. Text2SQL.} While Text2SQL generally yields the highest peak alpha (e.g., DeepSeek-Chat on TSLA), the PCG method offers specific advantages for ensuring consistency. For Gemini-2.5-Flash, PCG significantly improved outcomes on SPY, achieving the highest recorded ROC in the appendix (7.087). However, PCG proved less stable for DeepSeek-Chat, which regressed to negative profits on TSLA (-458.26) under this paradigm. This suggests a model-specific preference: reasoning-heavy models like Gemini benefit from the step-by-step decomposition of PCG, whereas models with strong raw instruction-following capabilities like DeepSeek-Chat thrive under the direct constraints of Text2SQL.

\begin{table*}[h]
  \centering
  \scriptsize
  \setlength{\tabcolsep}{1.8pt} 
  \caption{Query execution validity (VR), strategy-OK rate (SR), efficiency (Eff.), and average row count (Rows) across five underlyings.}
  \label{tab:full_metrics_21cols_standard}

  \adjustbox{max width=\textwidth}{%
  \begin{tabular}{lcccccccccccccccccccc}
    \toprule
    \multirow{2}{*}{\textbf{Model}} 
    & \multicolumn{4}{c}{\textbf{AAPL}} 
    & \multicolumn{4}{c}{\textbf{GOOG}} 
    & \multicolumn{4}{c}{\textbf{NVDA}} 
    & \multicolumn{4}{c}{\textbf{SPY}} 
    & \multicolumn{4}{c}{\textbf{TSLA}} \\
    \cmidrule(lr){2-5} \cmidrule(lr){6-9} \cmidrule(lr){10-13} \cmidrule(lr){14-17} \cmidrule(l){18-21}
    & \textbf{VR} & \textbf{SR} & \textbf{Eff.} & \textbf{Rows}
    & \textbf{VR} & \textbf{SR} & \textbf{Eff.} & \textbf{Rows}
    & \textbf{VR} & \textbf{SR} & \textbf{Eff.} & \textbf{Rows}
    & \textbf{VR} & \textbf{SR} & \textbf{Eff.} & \textbf{Rows}
    & \textbf{VR} & \textbf{SR} & \textbf{Eff.} & \textbf{Rows} \\
    \midrule
    DeepSeek-Chat    & 0.975 & 0.923 & 0.718 & 9.59 & 0.900 & 0.833 & 0.733 & 7.64 & 0.825 & \textbf{0.727} & 0.697 & 9.67 & 0.750 & 0.667 & 0.660 & 8.67 & 0.850 & \textbf{0.794} & 0.706 & 8.85 \\
    DeepSeek-Coder   & 0.750 & 0.833 & 0.773 & 9.70 & 0.650 & 0.769 & \textbf{0.754} & 8.65 & 0.575 & 0.478 & \textbf{0.748} & 7.87 & 0.675 & 0.481 & \textbf{0.763} & 9.33 & 0.650 & 0.692 & 0.738 & 9.39 \\
    Gemini 2.5 Flash & 0.900 & \textbf{1.000} & 0.717 & 9.53 & 0.875 & 0.771 & 0.663 & 8.03 & \textbf{0.950} & 0.632 & 0.658 & 9.18 & 0.875 & 0.600 & 0.669 & 10.0 & 0.900 & 0.694 & 0.672 & 9.94 \\
    GPT-4.1          & \textbf{1.000} & 0.950 & 0.670 & 9.50 & \textbf{1.000} & \textbf{0.850} & 0.670 & 8.38 & 0.875 & 0.657 & 0.680 & 9.20 & \textbf{0.900} & 0.611 & 0.639 & 10.0 & 0.875 & 0.771 & 0.703 & 9.63 \\
    GPT-4.1 Mini     & 0.975 & 0.949 & 0.769 & 9.74 & 0.975 & 0.795 & 0.703 & 8.05 & 0.925 & 0.486 & 0.697 & 8.78 & \textbf{0.900} & 0.556 & 0.722 & 9.92 & \textbf{0.950} & 0.658 & 0.726 & 9.71 \\
    Llama 3.1 8B     & 0.950 & 0.895 & 0.758 & 9.05 & 0.925 & 0.730 & 0.686 & 8.14 & 0.900 & 0.306 & 0.722 & 8.83 & 0.850 & 0.441 & 0.718 & 9.79 & 0.900 & 0.472 & 0.622 & 9.42 \\
    Qwen-Coder       & 0.850 & \textbf{0.971} & 0.741 & 9.18 & 0.850 & 0.735 & 0.641 & 8.00 & 0.675 & 0.593 & 0.674 & 8.67 & 0.725 & 0.621 & 0.703 & 9.24 & 0.625 & 0.680 & 0.616 & 9.76 \\
    Qwen3-4B         & 0.875 & 0.943 & \textbf{0.777} & 9.86 & 0.800 & 0.719 & 0.731 & 8.41 & 0.650 & 0.423 & 0.731 & 9.42 & 0.650 & 0.423 & 0.692 & 10.0 & 0.625 & 0.680 & \textbf{0.744} & 9.64 \\
    Qwen3-8B         & 0.850 & \textbf{0.971} & 0.671 & 9.38 & 0.850 & 0.824 & 0.659 & 8.50 & 0.775 & 0.677 & 0.600 & 8.45 & 0.775 & \textbf{0.677} & 0.677 & 8.77 & 0.725 & 0.793 & 0.600 & 9.10 \\
    \bottomrule
  \end{tabular}
  }
\end{table*}

\begin{table*}[h]
  \centering
  \scriptsize
  \setlength{\tabcolsep}{1.5pt} 
  \caption{Performance metrics across five underlyings: Win Rate (WR), Risk90, Profit, and ROC. (Top}
  \label{tab:model_performance_comparison_top}

  \adjustbox{max width=\textwidth}{%
  \begin{tabular}{lcccccccccccccccccccc}
    \toprule
    \multirow{2}{*}{\textbf{Model}} 
    & \multicolumn{4}{c}{\textbf{AAPL}} 
    & \multicolumn{4}{c}{\textbf{GOOG}} 
    & \multicolumn{4}{c}{\textbf{NVDA}} 
    & \multicolumn{4}{c}{\textbf{SPY}} 
    & \multicolumn{4}{c}{\textbf{TSLA}} \\
    \cmidrule(lr){2-5} \cmidrule(lr){6-9} \cmidrule(lr){10-13} \cmidrule(lr){14-17} \cmidrule(l){18-21}
    & \textbf{WR} & \textbf{Risk90} & \textbf{Profit} & \textbf{ROC}
    & \textbf{WR} & \textbf{Risk90} & \textbf{Profit} & \textbf{ROC}
    & \textbf{WR} & \textbf{Risk90} & \textbf{Profit} & \textbf{ROC}
    & \textbf{WR} & \textbf{Risk90} & \textbf{Profit} & \textbf{ROC}
    & \textbf{WR} & \textbf{Risk90} & \textbf{Profit} & \textbf{ROC} \\
    \midrule
    DeepSeek-Chat    & 0.615 & \textbf{0.103} & 209.1 & 0.207 & 0.400 & 0.229 & 22.1  & 0.000 & \textbf{0.697} & 0.152 & \textbf{481.3} & \textbf{0.686} & 0.531 & 0.406 & 104.8 & 0.358 & \textbf{0.800} & \textbf{0.114} & \textbf{984.0} & 0.529 \\
    DeepSeek-Coder   & 0.467 & 0.300 & 165.6 & 0.016 & 0.423 & 0.269 & 32.7  & 0.074 & 0.478 & 0.261 & 216.7 & 0.011 & \textbf{0.577} & 0.346 & 309.0 & 0.189 & 0.577 & 0.192 & 793.7 & 0.402 \\
    Gemini-2.5-Flash & \textbf{0.657} & 0.114 & 191.4 & 0.185 & 0.438 & 0.188 & -23.5 & 0.519 & 0.579 & 0.237 & 456.6 & 0.183 & 0.559 & 0.382 & \textbf{552.4} & \textbf{0.860} & 0.514 & 0.200 & 444.2 & 0.317 \\
    GPT-4.1          & 0.487 & 0.308 & 109.4 & -0.072 & 0.474 & 0.263 & 47.0  & -0.216 & 0.417 & 0.389 & 223.7 & 0.065 & 0.429 & 0.486 & 272.5 & 0.105 & 0.571 & 0.257 & 684.9 & 0.277 \\
    GPT-4.1-Mini     & 0.436 & 0.308 & 75.5  & -0.229 & 0.421 & 0.237 & 14.6  & \textbf{0.888} & 0.500 & 0.421 & 93.1  & 0.280 & 0.545 & 0.333 & 304.2 & 0.461 & 0.553 & 0.237 & 381.4 & 0.587 \\
    Llama-3.1-8B     & 0.472 & 0.250 & 50.6  & 0.022 & 0.368 & 0.184 & -144.8 & -0.157 & 0.417 & \textbf{0.083} & -9.0  & 0.014 & 0.353 & 0.324 & 4.8   & 0.237 & 0.471 & 0.206 & 117.7 & 0.305 \\
    Qwen-Coder       & 0.485 & 0.212 & 161.9 & 0.205 & 0.394 & 0.152 & 23.8  & -0.074 & 0.429 & 0.179 & -129.5 & 0.182 & 0.464 & 0.321 & 186.0 & -0.709 & 0.552 & 0.241 & 522.4 & 0.321 \\
    Qwen3-4B         & 0.541 & 0.108 & \textbf{302.2} & 0.112 & 0.300 & 0.267 & -48.8 & -0.240 & 0.346 & 0.231 & -36.7 & -0.005 & 0.417 & \textbf{0.250} & 164.2 & -0.121 & 0.385 & 0.269 & 284.3 & 0.054 \\
    Qwen3-8B         & 0.583 & 0.222 & 217.8 & 0.139 & \textbf{0.484} & \textbf{0.129} & -120.1 & -0.052 & 0.556 & 0.222 & 264.7 & 0.274 & 0.483 & 0.517 & 34.8  & 0.456 & 0.621 & 0.276 & 479.9 & \textbf{1.296} \\
    \bottomrule
  \end{tabular}
  }
\end{table*}

\section{Case Study}
\label{appendix:case_study}

We present several case studies in Table~\ref{tab:case_study_visuals_long} to illustrate the strategic advantages of our method over the baselines. In particular, we focus on two representative objectives: income enhancement and portfolio delta hedging. The results show that our approach consistently achieves lower tracking error, better-controlled risk exposure, and higher profit potential.

Compared with baseline methods, our approach demonstrates a clear advantage in accurately translating hedging intent into effective option strategies. As shown in Cases 2, 4, and 6, our method exhibits more precise inverse tracking of the underlying assets for hedging purposes. This capability is especially valuable for portfolio protection during periods of market stress, such as the market panic observed in April 2025. The improved hedging performance mainly stems from a better understanding of delta exposure and the ability of OQL to identify the most suitable hedging structures.

In addition, Cases 1 and 3 highlight the effectiveness of our method in income enhancement scenarios using spread strategies. We observe that strategies generated by OQL maintain stable tracking behavior relative to the underlying assets while delivering enhanced income. This stability is largely due to OQL’s ability to directly leverage real-time option chain information, enabling the selection of well-balanced and market-consistent option combinations.

\begin{longtable}{ 
    >{\centering\arraybackslash}m{0.05\linewidth}  
    m{0.40\linewidth}                                
    m{0.50\linewidth}                                
}
\caption{Qualitative case studies of intent-to-strategy translation} 
\label{tab:case_study_visuals_long} \\

\toprule
\textbf{ID} & \textbf{Description (Title \& Intent)} & \textbf{Result (Visual \& Comment)} \\
\midrule
\endfirsthead

\caption[]{Qualitative case studies (Continued)}\\
\toprule
\textbf{ID} & \textbf{Description (Title \& Intent)} & \textbf{Result (Visual \& Comment)} \\
\midrule
\endhead

\midrule
\multicolumn{3}{r}{\textit{Continued on next page...}} \\
\endfoot

\bottomrule
\endlastfoot

1 & 
\textbf{Moderately Bullish Call Ratio Spread} \newline \newline
\textbf{User Intent:} \newline
``I'm bullish but cautious at these levels (350). Buy one ATM call and sell two OTM calls at 380. I want to profit from a slow drift higher, not a spike.'' 
& 
\includegraphics[width=\linewidth, height=4cm, keepaspectratio]{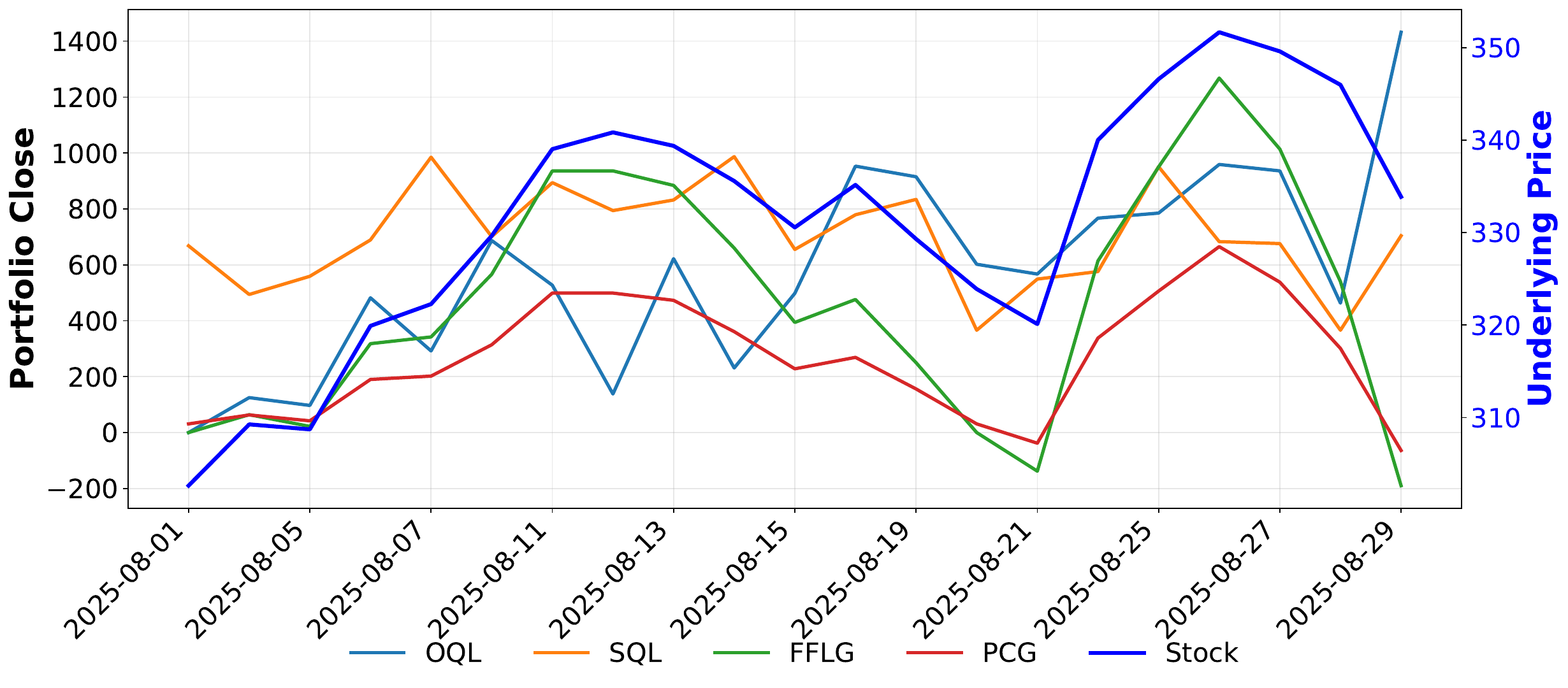} \newline
\small \textbf{Comment:} The model correctly interpreted the intent as a bullish call ratio spread, capturing upside from a gradual price increase while limiting gains under sharp rallies. The OQL strategy maintained positive returns during the underlying asset's volatile periods without experiencing significant drawdowns.
\\ \midrule

2 & 
\textbf{Delta-Neutral Iron Condor} \newline \newline
\textbf{User Intent:} \newline
``NVDA is range bound 160-180 (Sep). Construct a Delta Neutral Iron Condor. Harvest theta.'' 
& 
\includegraphics[width=\linewidth, height=4cm, keepaspectratio]{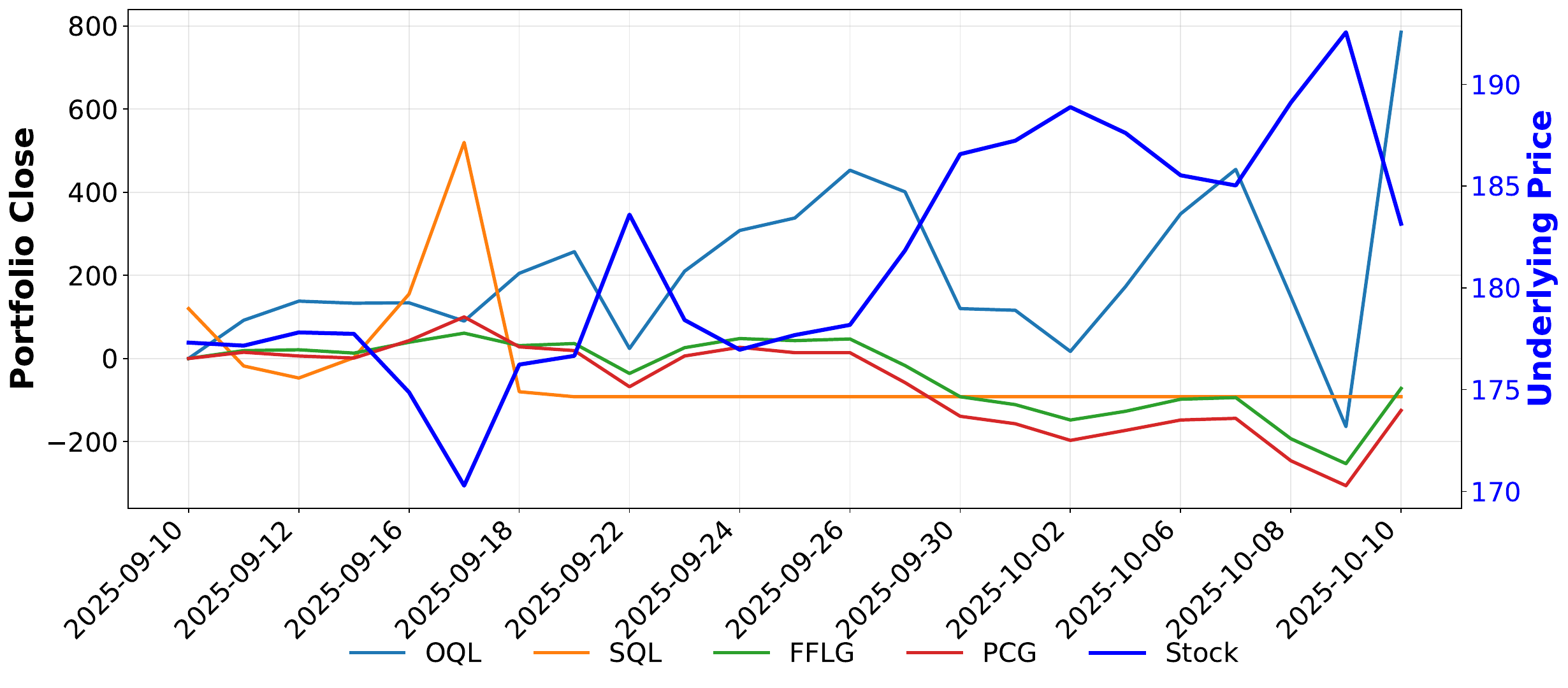} \newline
\small \textbf{Comment:} The model successfully mapped the range-bound and income-seeking intent to a delta-neutral iron condor, prioritizing theta decay while maintaining balanced directional exposure. The OQL strategy demonstrates staggering explosive power for inverse profits during sharp declines in the underlying asset.
\\ \midrule

3 & 
\textbf{Support-Driven Put Spreads} \newline \newline
\textbf{User Intent:} \newline
``The trendline support at 200 is holding beautifully. I'm willing to bet my house it doesn't drop below 195. Sell aggressive put spreads to finance a long position.'' 
& 
\includegraphics[width=\linewidth, height=4cm, keepaspectratio]{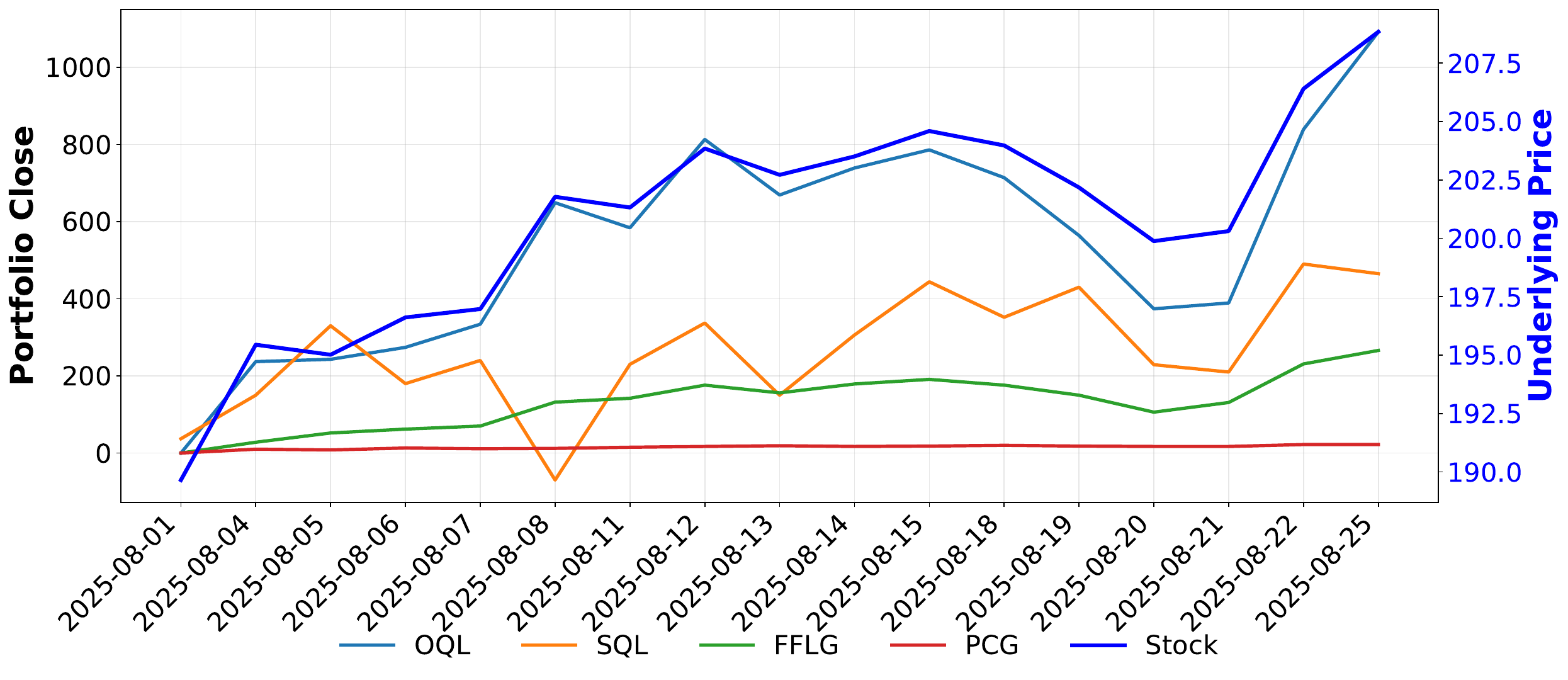} \newline
\small \textbf{Comment:} The model correctly interpreted the intent as a support-driven bullish strategy and mapped it to a call ratio spread (or equivalent bullish structure) that benefits from a gradual upside move while controlling downside risk. The OQL strategy perfectly synchronizes with the underlying asset's upward momentum.
\\ \midrule

4 & 
\textbf{Theta Harvest (Consolidation)} \newline \newline
\textbf{User Intent:} \newline
``Consolidation phase between 200 and 220. IV is still rich. Sell an Iron Condor to harvest theta as we chop sideways.'' 
& 
\includegraphics[width=\linewidth, height=4cm, keepaspectratio]{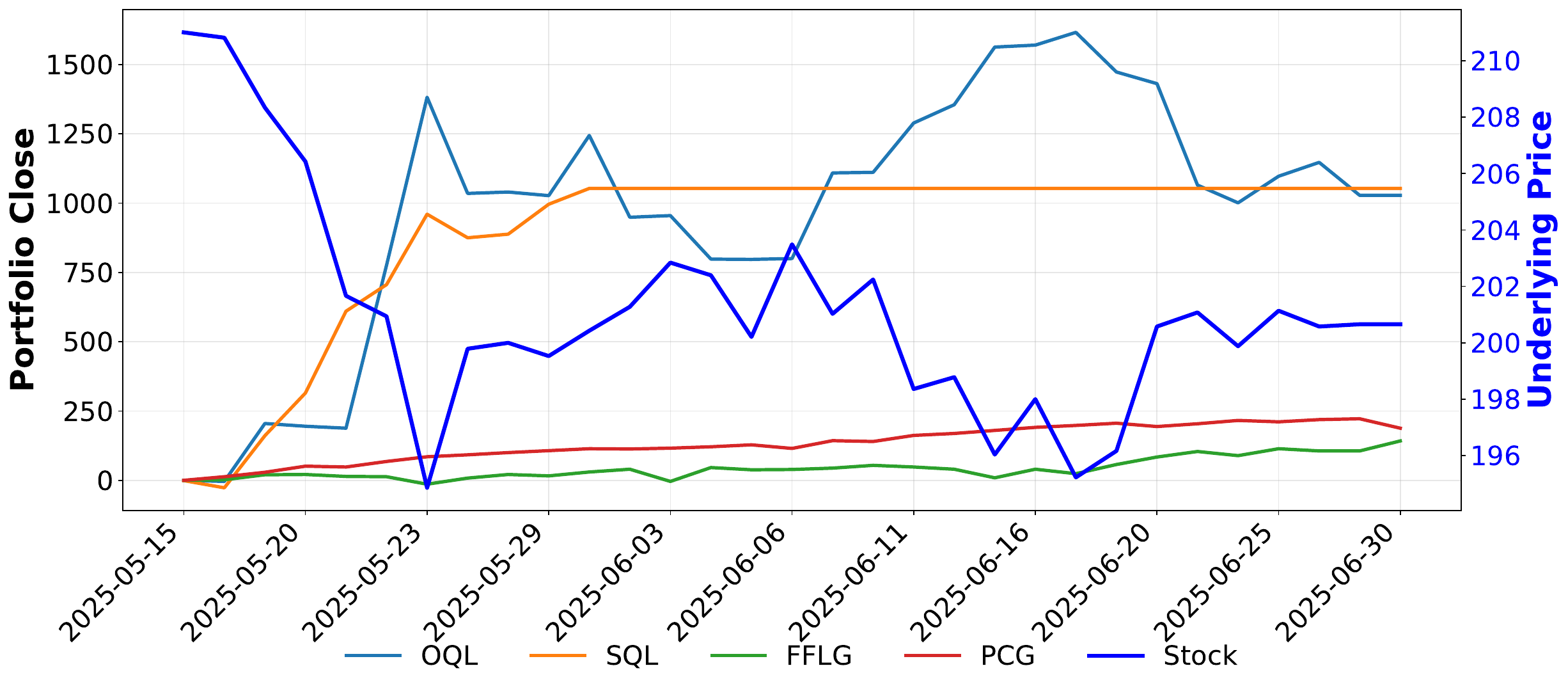} \newline
\small \textbf{Comment:} The model successfully translated the range-bound and volatility-rich intent into a delta-neutral iron condor, explicitly optimizing for theta decay. The OQL strategy achieves rapid portfolio value recovery and maintains a consistent upward trajectory.
\\ \midrule

5 & 
\textbf{Trendline Support (Variant)} \newline \newline
\textbf{User Intent:} \newline
``It's accelerating downside! 180 is gone. Target is 160. Get me a vertical put spread for the next two weeks to maximize ROI on this crash.'' 
& 
\includegraphics[width=\linewidth, height=4cm, keepaspectratio]{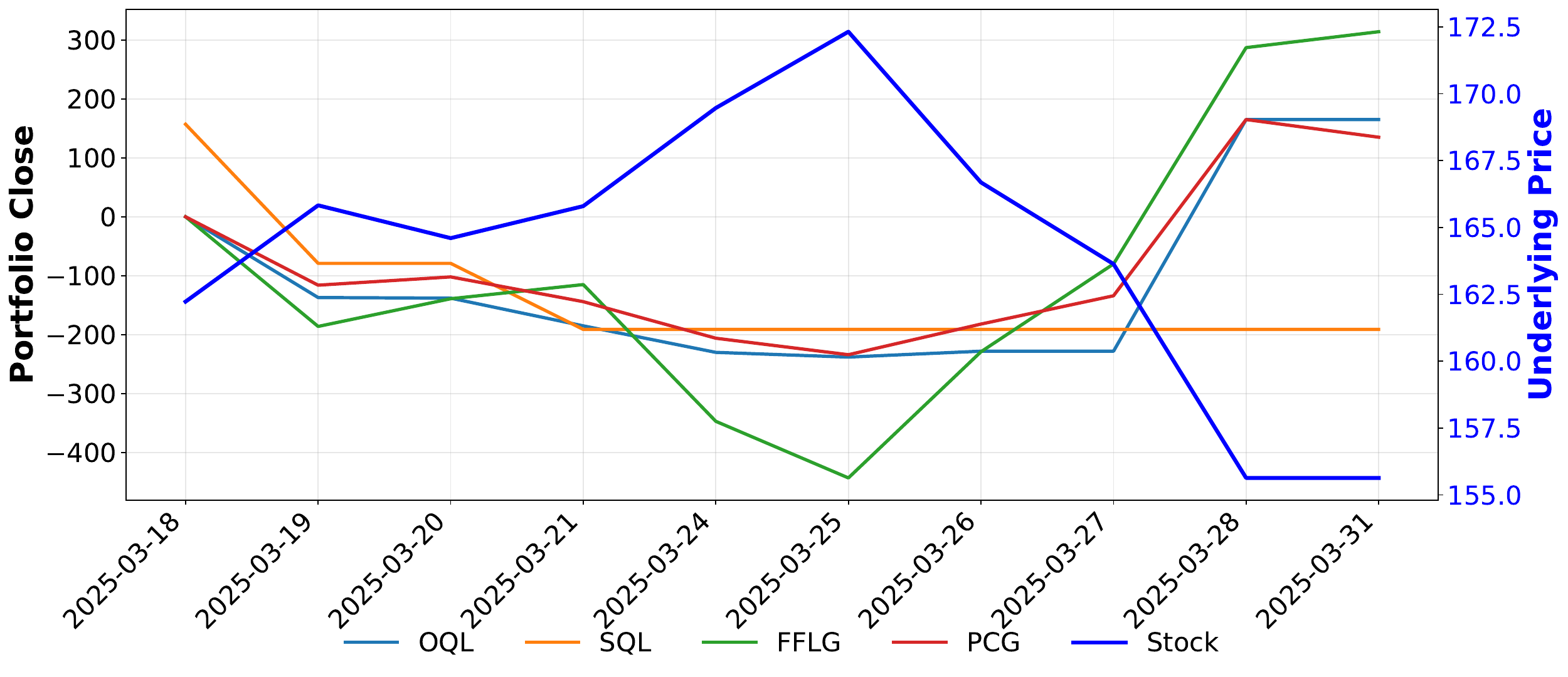} \newline
\small \textbf{Comment:} The model interpreted the intent as a bearish momentum strategy. The OQL and PCG strategies exhibit similar behavior, achieving rapid growth in returns during sharp declines in stock prices. In contrast, FFLG is subject to extremely massive drawdowns.
\\ \midrule

6 & 
\textbf{Volatility Rich Condor} \newline \newline
\textbf{User Intent:} \newline
``Panic selling at 100! Volume is insane. I don't know where the bottom is. Buy a Straddle. 30 days.'' 
& 
\includegraphics[width=\linewidth, height=4cm, keepaspectratio]{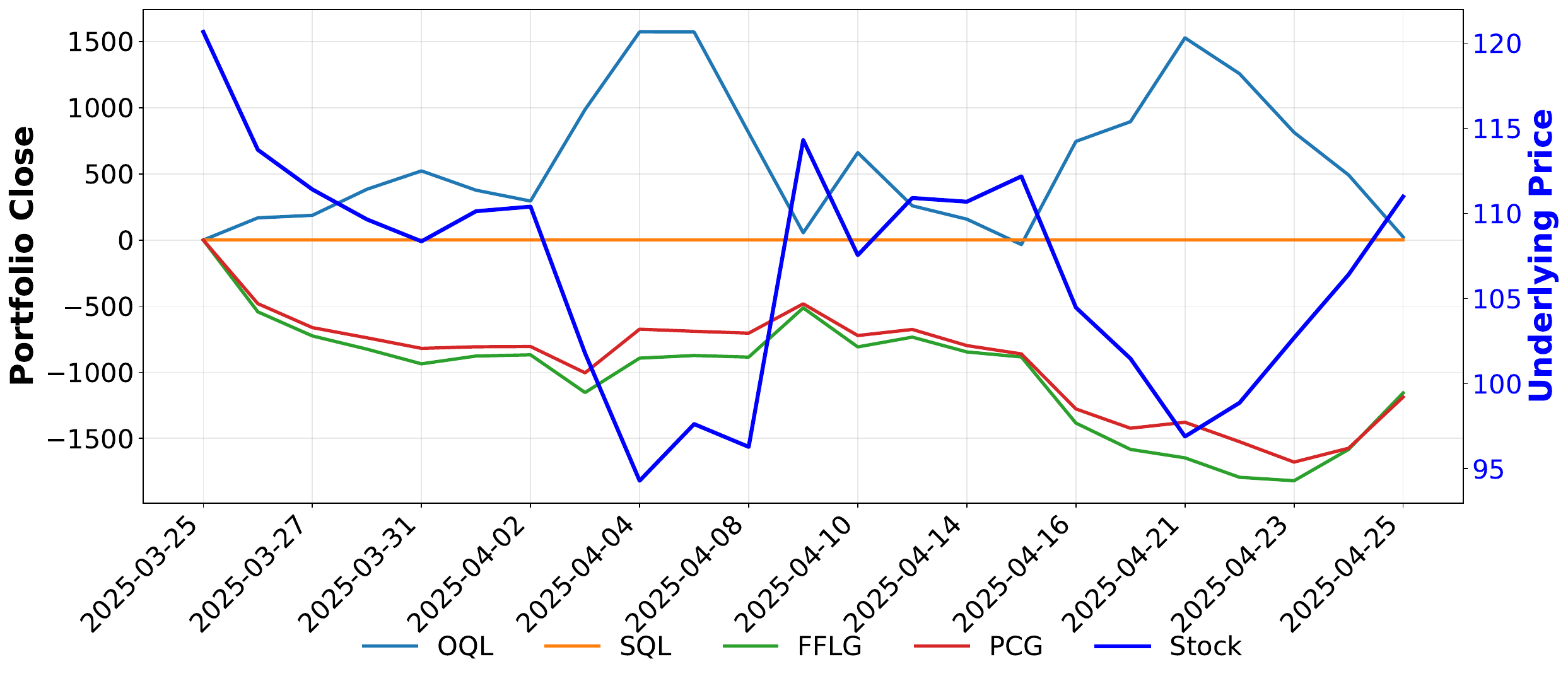} \newline
\small \textbf{Comment:} The model translated the intent into a high-volatility Straddle strategy. The OQL strategy demonstrates strong inverse profit characteristics. In contrast, the FFLG and PCG strategies suffer from prolonged negative returns, while SQL remains stuck at zero returns for the long term.
\\

\end{longtable}



\end{document}